\documentclass{article}

\usepackage{microtype}
\usepackage{graphicx}
\usepackage{subcaption}
\usepackage{booktabs} 
\usepackage{multirow}

\usepackage{hyperref}

\usepackage[preprint]{icml2026}

\usepackage{amsmath}
\usepackage{amssymb}
\usepackage{mathtools}
\usepackage{amsthm}
\usepackage{color, soul}
\usepackage[most,breakable]{tcolorbox}
\usepackage{xcolor}         
\usepackage{wrapfig}
\usepackage{amsthm}
\usepackage{tcolorbox}
\usepackage{siunitx}
\usepackage{threeparttable}
\usepackage{caption}
\usepackage{colortbl}
\usepackage{graphicx}
\usepackage{amssymb}
\usepackage{amsmath}
\usepackage{subcaption}

\usepackage[capitalize,noabbrev]{cleveref}

\theoremstyle{plain}
\newtheorem{theorem}{Theorem}[section]

\theoremstyle{definition}

\theoremstyle{remark}
\newtheorem{remark}[theorem]{Remark}

\newtcolorbox{mytheorem}{
  colback=gray!5, 
  colframe=gray!80, 
  boxrule=0.5pt, 
  arc=4pt, 
  left=4pt, 
  right=4pt, 
  top=4pt, 
  bottom=4pt, 
}

\newcommand{\hide}[1]{}

\usepackage[textsize=tiny]{todonotes}

\icmltitlerunning{Exploring Accuracy Law for Deep Time Series Forecasters: An Empirical Study}

\begin{document}

\twocolumn[

  \icmltitle{Exploring Accuracy Law for Deep Time Series Forecasters: An Empirical Study}



  \icmlsetsymbol{equal}{*}

  \begin{icmlauthorlist}
    \icmlauthor{Yuxuan Wang}{equal,thu}
    \icmlauthor{Haixu Wu}{equal,thu}
    \icmlauthor{Yuezhou Ma}{thu}
    \icmlauthor{Yuchen Fang}{byte}
    \icmlauthor{Ziyi Zhang}{byte}
    \icmlauthor{Yong Liu}{thu} \\
    \icmlauthor{Shiyu Wang}{byte}
    \icmlauthor{Zhou Ye}{byte}
    \icmlauthor{Yang Xiang}{byte}
    \icmlauthor{Jianmin Wang}{thu}
    \icmlauthor{Mingsheng Long}{thu}
  \end{icmlauthorlist}

  \icmlaffiliation{thu}{School of Software, Tsinghua University. Yuxuan Wang $<$wangyuxu22@mails.tsinghua.edu.cn$>$.}
  \icmlaffiliation{byte}{ByteDance China}

  \icmlcorrespondingauthor{Mingsheng Long}{mingsheng@tsinghua.edu.cn}

  \icmlkeywords{time series}

  \vskip 0.3in
]



\printAffiliationsAndNotice{\icmlEqualContribution}

\begin{abstract}
Deep time series forecasting has emerged as a rapidly growing field in recent years. Despite the exponential growth of community interests, progress on standard benchmarks is often limited to marginal improvements. A common consensus of the community is that time series forecasting inherently faces a non-zero error lower bound due to its partially observable and uncertain nature. However, a fundamental question arises: how to estimate the performance upper bound of deep time series forecasters? We delve into univariate time series forecasting, a prevalent forecasting paradigm spanning traditional statistical models to advanced time series foundation models. Going beyond classical series-wise predictability metrics, we realize that the forecasting performance is highly related to window-wise properties due to the sequence-to-sequence forecasting paradigm of deep time series models and introduce a quantitative measurement of window-wise pattern complexity. Through rigorous statistical analyses over more than 4700 newly trained deep forecasting models, we discover a consistent empirical relationship between the minimum attainable forecasting error of deep models and the complexity of window-wise series patterns, which is termed the \emph{accuracy law}. We further demonstrate that this empirical finding successfully guides us to identify saturated tasks from widely used benchmarks and derive an effective training strategy for time series foundation models, offering valuable insights for future research.
\end{abstract}

\vspace{-20pt}
\section{Introduction}
\begin{figure*}[t]
    \vspace{-5pt}
    \centering
    \includegraphics[width=\linewidth]{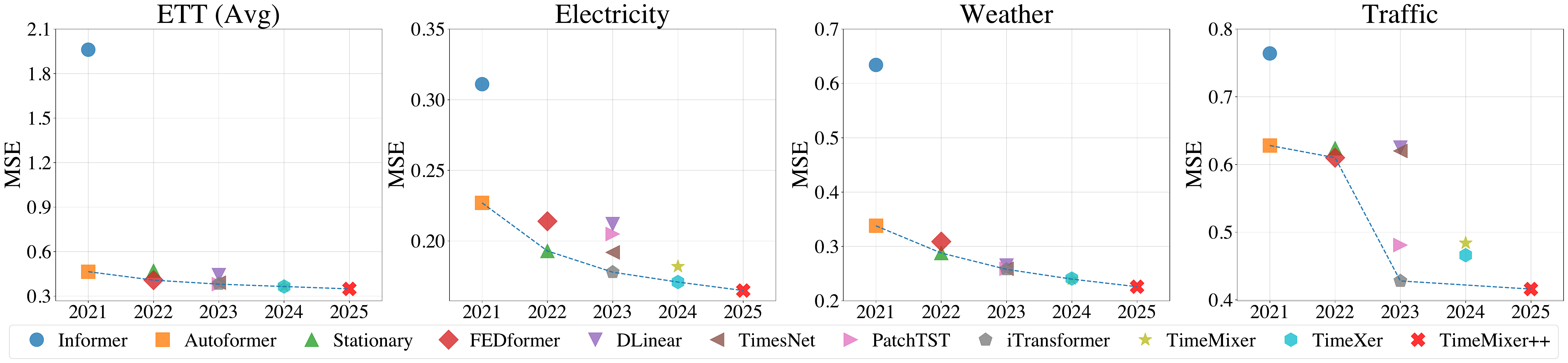}
    \vspace{-15pt}
    \caption{The performance change of deep time series forecasting in the past five years on well-established benchmarks. We record the MSE averaged from four widely used forecasting horizons.}
    \label{fig:model_performance}
    \vspace{-10pt}
\end{figure*}

In recent years, a significant number of deep learning models have been introduced for time series forecasting \cite{zhou2021informer,wu2021autoformer,wang2024deep}, which are equipped with elaborative architectures and meticulously crafted designs, demonstrating notable performance across diverse real-world forecasting scenarios, including finance \cite{gao2023stockformer, wang2024news}, transportation \cite{yu2017spatio, guo2019attention}, and meteorology \cite{wu2023interpretable}. Despite these advancements, we notice that performance improvements on widely used benchmarks have become increasingly marginal. As presented in Figure \ref{fig:model_performance}, the performance gains achieved by newly proposed deep forecasting models on four standard benchmarks have slowed significantly over the past three years. For instance, on the ETT benchmark \cite{zhou2021informer}, the relative forecasting performance improvements exhibited a  downward trend from 2022 to 2025, with values of 14.98\%, 7.77\%, 3.93\%, and 3.51\%, respectively. This stagnation has left the community increasingly confused about the research direction they are pursuing, as well as questioning the value of continuing to strive for minor improvements on these benchmarks.

Drawing inspiration from other tracks of the machine learning community like computer vision, we notice a fundamental difference between time series forecasting and these evergreen fields: \emph{does the field have a well-recognized and clear performance upper bound}? For instance, image recognition in computer vision \cite{deng2009imagenet}, a popular and long-standing area since 2000, has a quantifiable and widely accepted goal, that is, to achieve 100\% accuracy. Although the 100\% accuracy may be practically unattainable due to labeling noise and inherent ambiguity in some images, these well-established targets provide a concrete reference of model capability and ensure the community is clear on whether one task or benchmark has been well solved~\cite{krizhevsky2012imagenet, simonyan2014very, he2016deep}. Unfortunately, there is no such clear goal for time series forecasting. Given the partially observable and inherently uncertain nature of time series, achieving zero prediction error is fundamentally unattainable. More importantly, even human experts often struggle to define what constitutes the best possible prediction for a given time series, making this area suffer from saturated tasks and uncertain measures of actual model quality.

Beyond general time series forecasting, certain real-world applications exhibit domain-specific notions of a forecasting performance upper bound, often informed by decades of expert experience, such as weather forecasting \cite{wu2023interpretable} and financial quantification \cite{yang2020qlib}. However, such domain knowledge relies on human experts and is limited to specific areas, which cannot serve as a general rule for diverse and complex forecasting tasks studied by the research community. Additionally, expert knowledge may result in a wrong judgment of the bottleneck due to personal bias or human limitations, which can be potentially broken by data-driven methodology \cite{bi2023accurate}. These observations highlight the need for a \emph{performance upper bound} for data-driven models to better contextualize real progresses in deep time series forecasting.

Inspired by statistical estimation theory \cite{kay1993fundamentals, cramer1999mathematical}, which rigorously defines the fundamental limits on the precision of any estimator, we explore whether there exists a general and instructive performance upper bound for time series forecasting, which is determined by the inherent predictability of time series. Building on the observation that deep time series forecasters predominantly follow a sequence-to-sequence paradigm, we introduce a window-wise complexity metric to quantify the intrinsic complexity of temporal patterns within a time series. Through thousands of empirical experiments and rigorous statistical tests, we discovered the \emph{accuracy law} for deep time series forecasters. Specifically, we observe a significant exponential relation  between the window-wise pattern complexity of a given series and its minimum forecasting error achieved by state-of-the-art deep models. This accuracy law provides empirical insights
for understanding predictability in the context of deep time series forecasting and sheds light on future research directions. Based on the proposed accuracy law, we successfully identify the saturation of tasks within widely used benchmarks, releasing the research community from diminishing returns on over-studied datasets. Besides, the proposed complexity measure can be used to derive a simple yet effective training strategy, which further improves the performance of time series foundation models. Our contributions are summarized:
\begin{itemize}
    \item We notice a fundamental issue in the community of deep time series forecasting, that is, the absence of a general lower bound for deep time series models.
    \item Through extensive experiments across diverse datasets and model architectures, we empirically formulate the \emph{accuracy law} for deep time series forecasters. This law uncovers that the minimum forecasting error achieved by existing deep forecasters exhibits an exponential relationship with the window-wise pattern complexity of time series, which demonstrates favorable generalizability across diverse forecasting tasks.
    \item The proposed complexity measure and accuracy law provide valuable insights for the community, which can seamlessly support identifying saturated tasks and training time series foundation models.
\end{itemize}

\section{Related Works}

\subsection{Time Series Predictability}

Time series predictability \cite{box2015time} refers to the intrinsic degree to which future observations of a given series can be accurately inferred or estimated from past observations. It is a fundamental concept in time series analysis that provides an a priori understanding of the best achievable complexity for a given time series, independent of the method. Consequently, accurately quantifying this property is a crucial prerequisite for evaluating the potential of the forecasting task and establishing a realistic performance benchmark. Over the past decades, researchers have developed various well-established statistical methods to quantify predictability \cite{pincus1991approximate,hamilton2020time}.

Traditional forecasting methodologies often rely on the assumption of stationarity \cite{granger2001spurious}. Therefore, assessing stationarity has become an important preliminary step in evaluating series predictability. The Augmented Dickey-Fuller (ADF) test \cite{elliott1992efficient} is a statistical method for determining the time series stationarity. A smaller ADF-statistic, leading to the rejection of the null hypothesis, suggests a higher degree of stationarity, also potentially indicating higher predictability. Beyond stationarity, entropy-based measures offer a more concise way to quantify the complexity and predictability of a time series \cite{richman2000physiological}. ForeCA \cite{goerg2013forecastable} introduces a quantitative measure of forecastability based on spectral entropy of the series. This metric can be calculated by subtracting the entropy of the series' Fourier domain, where a higher value signifies superior predictability, also offering a model-agnostic insight. 
However, most of the existing predictability measures are designed for an entire series, which is not consistent with the sequence-to-sequence forecasting paradigm for deep models. Further, the measurement of predictability in the context of deep time series forecasting remains relatively underexplored.

\vspace{-5pt}
\subsection{Time Series Forecasting Benchmarks}

Evaluation benchmarks are fundamental in driving progress by providing researchers with clear and well-defined goals, which is acknowledged as one of the keys to ``the second half of AI.'' With the rapid growth of time series forecasting models, a series of benchmarks has emerged to meet the growing need for rigorous and comprehensive model evaluations. Informer \cite{zhou2021informer} introduced a standardized training and evaluation protocol for long-term time series forecasting, along with two widely used datasets, ETT and Electricity. Subsequent work, Autoformer~\cite{wu2021autoformer}, further expanded benchmark coverage to more diverse domains, including weather, health, and finance. Additional datasets, such as Solar-Energy~\cite{lai2018modeling} and traffic benchmarks (e.g., PEMS), are also routinely used in the following research~\cite{liu2022scinet,wang2024timemixer}.
More recently, the emergence of time series foundation models has raised new challenges for fair and comprehensive evaluation, particularly in zero-shot forecasting. In response, several large-scale benchmarks have been developed. \cite{ansari2024chronos} developed FEV, an open leaderboard dedicated to zero-shot forecasting. \cite{aksu2024gift} proposed GIFT-Eval, a framework featuring distinct pretraining and train/test components tailored for evaluating time series foundation models. 


\section{Accuracy Law}
As aforementioned, we attempt to build a {general and instructive} performance upper bound for deep time series forecasting, which is assumed to be relevant to time series predictability. To obtain a formal and statistically significant relationship, we start from a huge hypothesis space with a wide consideration of predictability and performance metrics, as well as relation types. However, since deep models typically employ a sequence-to-sequence forecasting paradigm, the classical series-wise predictability metrics fail in presenting a significant relation to final performance. Stemming from the deep forecasting paradigm, we newly propose a window-wise pattern complexity and finally establish an accuracy law through thousands of experiments and rigorous statistical analysis.

\subsection{Hypothesis Space}\label{sec:space}
Although the assumption of correspondence between time series predictability and the performance upper bound of deep forecasting seems reasonable, it is not easy to discover a general and significant law that can formalize the assumed correspondence. To narrow down the hypothesis space and isolate a significant relationship from intricate factors, our work is within the following scope:

\underline{\emph{(i) Univariate forecasting.}} In this paper, we only consider the univariate forecasting task. This scope can avoid the inherent performance trade-off in multivariate forecasting, allowing our statistical test to focus on temporal predictability. Besides, since most of the advanced deep time series models follow channel-independent training \cite{nie2022time} and time series foundation models are predominantly univariate \cite{shi2024time,liu2025sundial}, this scope will not damage the practicability of the conclusion.

\underline{\emph{(ii) Approximated performance upper bounded.}} Since the exact ``optimal'' forecast performance is unreachable, for each time series, we take the lowest error among several state-of-the-art forecasters as a surrogate. Despite this being only an approximation, the lowest value reflects the upper bound achieved by current research, which is already enough to identify which task falls beyond or behind the frontier, as well as to reflect the task's inherent difficulty. More importantly, taking the lowest error among various models also offers a model-agnostic perspective for analysis.

\begin{figure}[t]
  \begin{center}
    \includegraphics[width=0.4\textwidth]{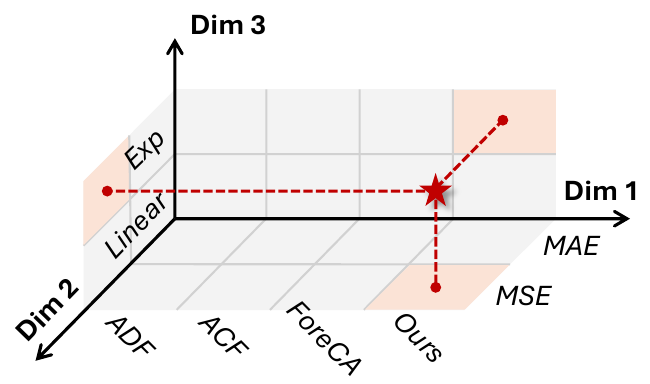}
  \end{center}
  \vspace{-10pt}
  \caption{\small{Hypothesis space considered in this paper. The discovered accuracy law is marked with a red star.}}\label{fig:hypothesis}
  \vspace{-20pt}
\end{figure}
Building up the above two basic settings, we examine a wide variety of performance and predictability metrics to characterize their underlying relationship (Fig.~\ref{fig:hypothesis}). The key factors can be organized in three dimensions:

\textbf{(i) Dim 1:} Predictability metric (\underline{\emph{4 choices}}): We consider three classical predictability indicators: Augmented Dickey-Fuller (ADF) test \cite{elliott1992efficient} statistic, ForeCA \cite{goerg2013forecastable} and the half-life of the autocorrelation function (ACF) \cite{ramsey1974characterization}, which have served as the foundation of time series analysis. We also proposed a window-wise complexity metric, which will be detailed later.

\textbf{(ii) Dim 2:} Performance metric (\underline{\emph{2 choices}}): We adopt the widely used metrics, Mean Absolute Error (MAE) and Mean Squared Error (MSE), to quantify the forecasting error.

\textbf{(iii) Dim 3:} Relation formalization (\underline{\emph{2 choices}}): 
Guided by the classical \emph{principle of Occam's razor}, we only consider statistical tests on exponential\footnote{The exponential relation test is conducted via the linearity w.r.t.~Log Mean Squared Error (LogMSE) and Log Mean Absolute Error (LogMAE), which are defined as $\log (\operatorname{MSE}+1),\log(\operatorname{MAE}+1)$ to maintain the theoretical intuition that \emph{the model performance metric should be zero when predictability metric is zero}.}  and linear relations, which are simple but sufficient.

After training 4700 time series forecasting models as statistical samples and conducting rigorous statistical tests for the above 16 combined possible hypotheses, we have only observed one significant exponential relationship between our newly proposed complexity and MSE, which is named as \emph{accuracy law for deep time series forecasting.} This finding can be briefly described as follows.

\begin{tcolorbox}[colback=blue!2!white,leftrule=2.5mm,size=title]
\textbf{Accuracy law (Informal).} 
\textit{Within a certain interval of predictability, the lowest MSE of deep models exhibits a clear exponential relation with our proposed window-wise complexity.}
\end{tcolorbox}

Next, we present the definition of window-wise complexity and corresponding statistical results.

\subsection{Window-wise Pattern Complexity}

According to our experiments that will be detailed in Section \ref{sec:comparison}, all traditional series-wise predictability metrics \cite{elliott1992efficient,goerg2013forecastable} fail in demonstrating a value relationship. Thus, we attempt to present a new predictability metric tailored to the typical deep forecasting paradigm. 

\vspace{-5pt}
\paragraph{Insights from forecasting paradigm} Typically, deep models utilize the sequence-to-sequence forecasting paradigm, where the model is expected to predict\hide{the future} a future segment based on the latest segment of past observations \cite{zhou2021informer,wu2021autoformer,wang2024deep}. Specifically, given a time series $\mathbf{x}=\{\cdots,\mathbf{x}_t,\cdots\}$, its forecasting paradigms can be formalized as follow,
\begin{equation}\label{equ:paradigm}
\begin{aligned}
\text{Train:}& \max_{\theta}\sum\nolimits_{t}\mathbb{P}\big(\mathbf{x}_{t+1:t+F}|\mathbf{x}_{t-P:t},f_{\theta}\big), \\
\text{Inference:}&~\widehat{\mathbf{x}}_{t+1:t+F}=f_{\theta}(\mathbf{x}_{t-P:t}),
\end{aligned}
\end{equation}
where $f_{\theta}$ represents the forecasting model with parameter $\theta$. Here, $F$ and $P$ denote the lengths of the prediction horizon and past observation, respectively. The above paradigm motivates us that the predictability of forecasting models would be closely related to the window-wise property of time series rather than series-wise. Therefore, we present the following window-wise pattern complexity.

\vspace{-5pt}
\paragraph{Complexity definition} Inspired by the above analysis, for each time series, we consider the continuous time windows with length $(P+F)$, which can completely cover all the time points used in a single forecast. Further, to highlight the variation pattern in time series \cite{wu2023timesnet}, we transform the time window into the frequency domain using the Fast Fourier Transform (FFT) and extract its amplitude spectrum only, which represents the magnitudes of all frequency components and demonstrates the strength of various variation patterns. This transformation also allows our metric to mitigate the influence of temporal shifts that are recorded in the phase information of the Fourier domain. To sum up, the computation process can be formalized as:
\begin{equation}\label{equ:fft}
\begin{aligned}
    \{\mathbf{x}_{i:(i+P+F)}\}_{i} &= \operatorname{Split}(\mathbf{x}), \\
    \ \{\mathbf{A}_i\}_i &= \left\{\operatorname{Amp}\big(\operatorname{FFT}\left(\mathbf{x}_{i:(i+P+F)}\right)\big)\right\}_i,
\end{aligned}
\end{equation}
where $\operatorname{Amp}(\cdot)$ represents keeping the amplitude spectrum only and $\mathbf{A}_i\in\mathbb{R}^{P+F}$ denotes the extracted pattern of the $i$-th window. Based on this frequency domain representation, we define the {series pattern complexity} as the total variance of \emph{amplitude spectrum distribution}, which records the spread of the amplitude spectra distribution among all time windows and inherently describes the distribution diversity. Thus, the proposed series pattern complexity of time series $\mathbf{x}$ is defined as:
\begin{equation}\label{equ:definition}
\begin{aligned}
    \operatorname{Complexity}(\mathbf{x}) &= \operatorname{tr}(\operatorname{Cov}(\{\mathbf{A}_i\})) \\
    &= \frac{1}{N} \sum\nolimits_{1 \leq i\leq N}  ||\mathbf{A}_i-\bar{\mathbf{A}}||^2_2.
\end{aligned}
\end{equation}
Here, $\operatorname{tr}(\cdot)$ denotes the matrix trace and $\operatorname{Cov}(\cdot)$ denotes the covariance operator, where $N$ denotes the number of divided time windows and $\bar{\mathbf{A}}$ represents the sample mean of $\{\mathbf{A}_i\}$. The above complexity metric characterizes the intrinsic heterogeneity of series variations in every interested window, where a higher value suggests the series contains more distinct  patterns and is harder to predict\hide{, whereas a lower value points to greater homogeneity}.

\begin{figure}[h]
  \begin{center}
    \includegraphics[width=0.49\textwidth]{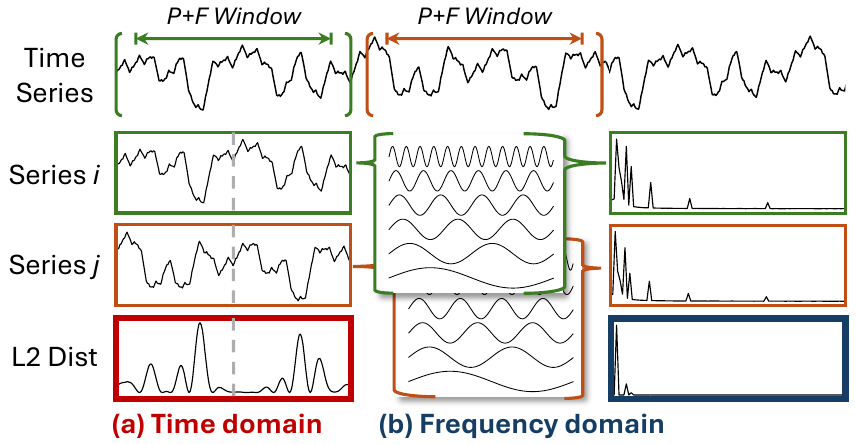}
  \end{center}
  \vspace{-5pt}
  \caption{\small{Illustration of time and frequency distance.}}\label{fig:understanding}
\end{figure}

\begin{figure*}[t]
    \centering
    \includegraphics[width=\linewidth]{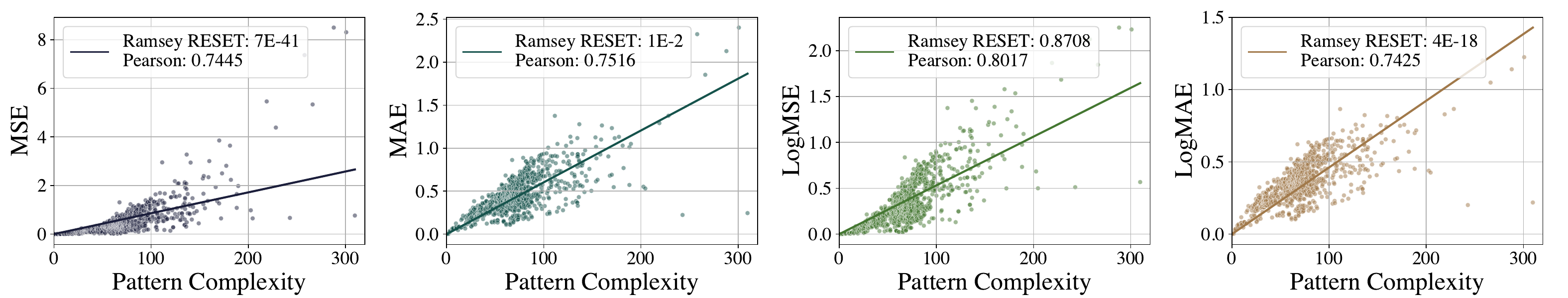}
    \vspace{-15pt}
    \caption{Experimental results from 940 series and 5 deep forecasters for the accuracy law, which are processed from 4,700 experiments. For each data point, its $x$-coordinate denotes the window-wise pattern complexity calculated by Eq.~\ref{equ:definition} and the $y$-coordinate is the lowest forecasting error achieved among five experimental models: DLinear, PatchTST, TimeMixer, xPatch, and TimeMixer++.}
    \label{fig:accuracy-law}
    \vspace{-10pt}
\end{figure*}
\vspace{-5pt}
\paragraph{Understanding window-wise pattern complexity}We notice that the \emph{transformation to Fourier domain} is key to the above definition, which also yields instructive insights for subsequent statistical tests. More empirical comparisons to prove its superiority can be found in the Section \ref{sec:comparison}.

\emph{(i) Joint distribution modeling.} As presented in Figure \ref{fig:understanding}, we consider both past observations and the future horizon within the divided time windows. A simple option is to directly compute variance in the original time domain. However, since the time domain L2 distance is computed independently for different timestamps, the time domain variance can not capture the whole covariance matrix, thereby neglecting the mutual information between the past and future. In contrast, due to the \emph{global nature of the frequency domain}, which we have carefully selected, our definition in Eq.~\ref{equ:definition} naturally captures the joint distribution of past and future. which can be interpreted as simultaneously characterizing both the initial state and the transition dynamics.

\emph{(ii) Temporal shift invariant.} Benefiting from our design in Eq.~\ref{equ:fft}, which only considers the amplitude spectrum in the frequency domain while ignoring the phase term, for any lag $\tau$, the series $\{\mathbf{x}_t\}$ and its $\tau$-lagged counterpart $\{\mathbf{x}_{t-\tau}\}$ exhibit the same complexity. This enables our metric to capture principal variation in time series and avoid noise, thereby better reflecting the inherent predictability.

\subsection{Statistical Test}\label{sec:exp_detail}

In our narrowed hypothesis space, we assume a fundamental exponential or linear relation between the time series predictability metric and its achievable optimal forecasting performance. To validate this, we conduct a systematic and comprehensive empirical evaluation of the relation between our newly defined window-wise pattern complexity (Eq.~\ref{equ:definition}) and the minimum forecasting error achieved by several state-of-the-art deep forecasters. The following is the detail of our statistical test.

\vspace{-10pt}
\paragraph{Experimental data} 
To ensure comprehensive validation, we leverage LOTSA \cite{woo2024unified}, a large-scale public archive of real-world univariate time series from nine domains. The dataset provides coverage of varied temporal dynamics commonly encountered in real-world forecasting tasks. To alleviate domain imbalance, we randomly sample 20 series from each dataset and discard those shorter than 5000 time steps to meet the training requirements of deep models, yielding 940 univariate time series in total. Each series is treated as an independent task, and models are trained on a series-by-series basis.

\vspace{-8pt}
\paragraph{Experimental models} To measure the achievable forecasting performance, we include five state-of-the-art deep time series forecasters, including PatchTST \citep{nie2022time}, DLinear \citep{zeng2023transformers}, TimeMixer \citep{wang2024timemixer}, and TimeMixer++ \citep{wang2024timemixer++}, and xPatch \cite{stitsyuk2025xpatch}. These models have demonstrated exceptional performance on various benchmarks and are widely recognized for their effectiveness in univariate time series modeling.

\vspace{-8pt}
\paragraph{Implementation Details} Following established conventions \citep{wang2024deep}, we split the entire time series into training, validation, and test sets with a ratio of 7:1:2 . Experimentally, we train and evaluate all five models on the selected 940 time series independently under the input-96-predict-96 setting. For each time series, we take the best performance achieved across five experimental models as one data point for subsequent statistical tests.

\vspace{-8pt}
\paragraph{Statistical analysis} As shown in Figure~\ref{fig:accuracy-law}, we identify a clear \emph{linear relation between the proposed window-wise pattern complexity and the forecasting LogMSE}. To obtain a rigorous and formal description of the linear relationship, we formalize this observation from two perspectives.

\emph{(i) Linear dependence test.} First, a linear regression analysis is conducted to quantify the dependence between pattern complexity and various evaluation metrics. Among them, LogMSE achieves the strongest linear association, with a Pearson correlation coefficient \cite{benesty2009pearson} of 
77.67\%, which is generally considered to indicate a \emph{strong positive linear correlation} in statistical domains.

\begin{figure*}[t]
    \vspace{-10pt}
    \centering
    \includegraphics[width=\linewidth]{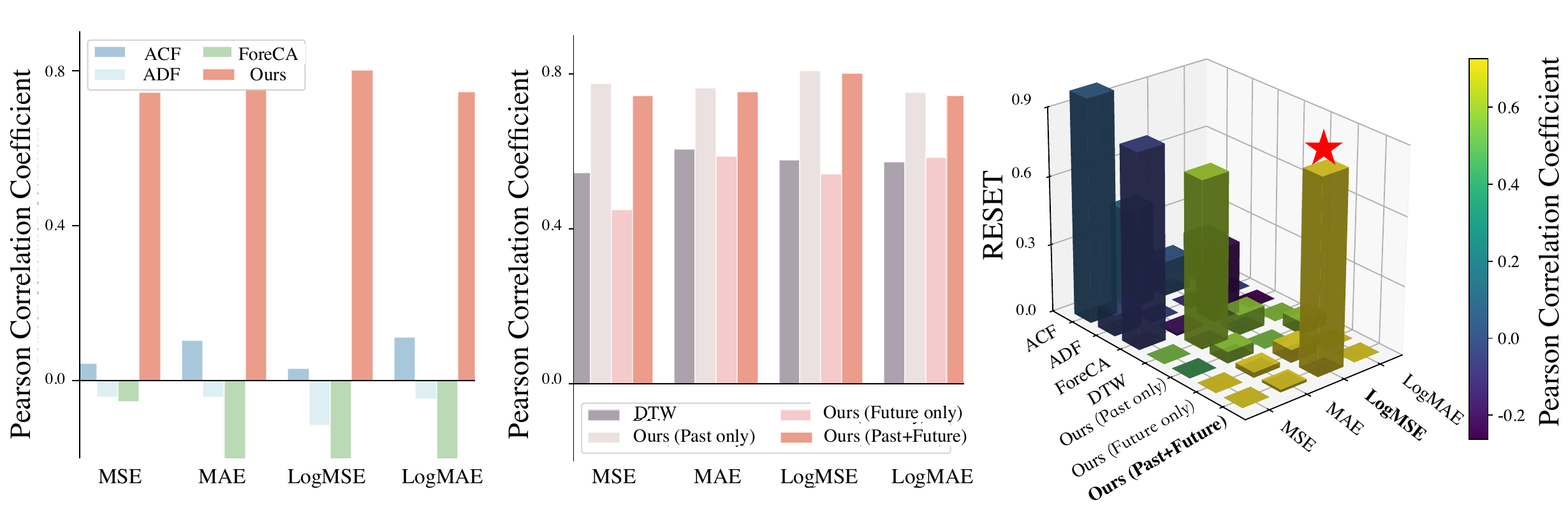}
    \vspace{-15pt}
    \caption{Comparison of our window-wise complexity with classical metrics (left) and other variations (middle), where Pearson coefficient between the predictability and performance is recorded. For the right part, the Ramsey RESET test is presented as the z-axis value; a higher bar indicates greater confidence in the linear relation; the brighter color refers to the higher Pearson coefficient.}
    \label{fig:statistical-results}
    \vspace{-10pt}
\end{figure*}

\emph{(ii) High-order dependence test.} Second, to avoid neglecting higher-order dependence, we further perform the Ramsey RESET test \citep{volkova2013research} on the second-order relation to examine whether higher-order terms are statistically required in precisely describing the relationship. Here, a lower p-value indicates greater confidence in high-order dependence. For LogMSE, the p-value is 0.85, well above the 0.05 significance threshold, indicating that introducing higher-order terms does not provide additional explanatory power. In contrast, the p-values for MAE, MSE and LogMAE are much smaller than 0.05, revealing the high-order dependence among these metrics and complexity.

\vspace{-5pt}
\paragraph{Summary} Based on the above results, we discovered a pure and confident exponential relationship between MSE and  complexity, formally defined in Eq.~\ref{equ:law}.

\vspace{-2pt}
\begin{tcolorbox}[colback=blue!2!white,leftrule=2.5mm,size=title]
\textbf{Accuracy law.} \textit
{There exists an interval of window-wise pattern complexity $(C_{\min}, C_{\max})$ such that, for any time series $\mathbf{x}$ with $\operatorname{Complexity}(\mathbf{x}) \in (C_{\min}, C_{\max})$, the minimum forecasting MSE achieved by all deep models admits a exponential relation with the complexity, which can be quantified as follows:}
\begin{equation}\label{equ:law}
   \mathrm{MSE} \approx  \mathrm{exp}\,({\alpha \cdot \operatorname{Complexity}(\mathbf{x})})-1,
\end{equation}
{\textit{where $\operatorname{Complexity}(\mathbf{x})$ represents our newly proposed pattern complexity measurement of time series $\mathbf{x}$. In our statistical test experiments above, $C_{\min}=0, C_{\max}=309, \alpha=0.0053$.}}
\end{tcolorbox}

\vspace{2pt}
This accuracy law provides an empirical framework for understanding time series predictability in the context of deep forecasting. Intuitively, time series with lower pattern complexity generally exhibit smaller forecasting errors, positioning them at the lower end of the fitted line and indicating better predictability. Importantly, the linear relation connecting these two regimes is non-trivial but has been rigorously validated through our statistical tests. Moreover, it is worth noting that the intercept of the accuracy law equals zero, implying that as the complexity of a time series approaches zero, the infinity of its forecasting MSE under deep models also vanishes, which aligns with the intuition of perfect predictability. Further, the accuracy law serves as both a predictive and prescriptive tool. By calculating the pattern complexity of time series, it becomes possible to estimate the expected forecasting error using this empirical relationship, providing a practical reference for performance evaluation and guidance for future model improvements.

\begin{remark}[The value of coefficient $\alpha$] The accuracy law in Eq.~\ref{equ:law} is tested under the input-96-predict-96 scenario, which means changing the forecasting setting may affect the fitted relationship. Thus, we also experiment on different lookback and forecasting settings, which can be found in Appendix~\ref{appdix:general_seq}. It can be observed that although the concrete value of $\alpha$ may vary, the experimental results demonstrate that the exponential relationship between MSE and pattern complexity is robust and general. It is really hard to identify a universal constant across various scenarios, and such a constant does not exist even in the scaling law of LLMs \cite{kaplan2020scaling,hoffmann2022training}.
\end{remark}

\section{Experiments}
To further demonstrate the discovered accuracy law's (Eq.~\ref{equ:law}) fundamentality and value to the deep forecasting research, this section will first make a comprehensive comparison with conventional predictability metrics and then showcase the insights for deep and time series foundation models.

\subsection{Overall Comparison}\label{sec:comparison}

As stated in Section \ref{sec:space}, we consider a huge hypothesis space to explore the accuracy law. Here, we present the statistical test on each hypothesis. Additionally, to examine the effectiveness of our proposed window-wise complexity, we also include the statistical test on some variants.

\vspace{-10pt}
\paragraph{Main results} From the statistical results in Figure \ref{fig:statistical-results}, we can obtain the following observations:

\begin{figure*}[t]
\vspace{-5pt}
    \centering
    \includegraphics[width=\linewidth]{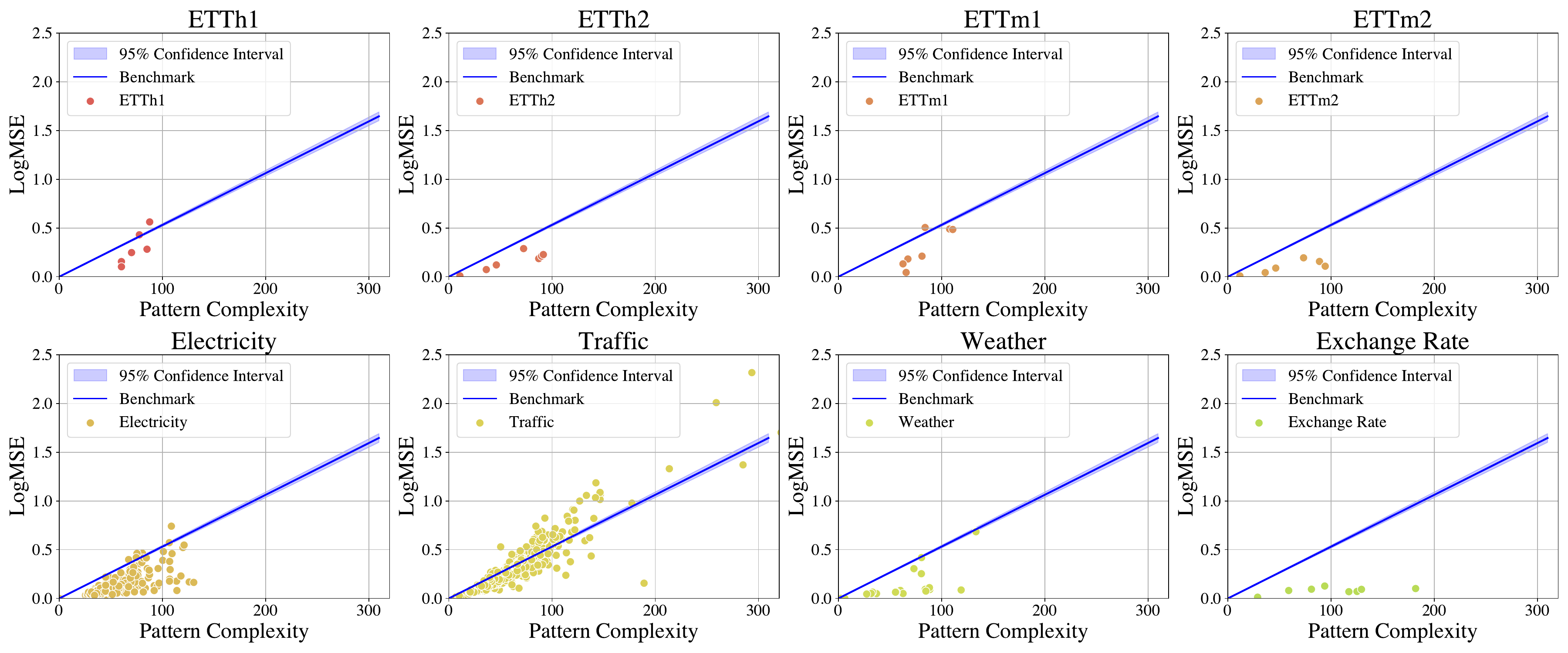}
    \vspace{-15pt}
    \caption{Analysis on widely used standard benchmarks. The scatter points represent individual time series from different benchmarks, and forecasting performance is measured by the minimum univariate forecasting error achieved by five tested state‑of‑the‑art deep models as described in Figure~\ref{fig:accuracy-law}. For a benchmark containing multiple series, if most series exhibit forecasting errors substantially below the estimation, we can consider this benchmark as saturated. Detailed results are listed in Table \ref{fig:app-box-vis}.}
    \label{fig:saturation}
\vspace{-10pt}
\end{figure*}

\emph{(i) Comparison with conventional series-wise indicators.} As aforementioned, the Augmented Dickey-Fuller (ADF) test \cite{elliott1992efficient} statistic, and the ForeCA \cite{goerg2013forecastable} and the half-life of the autocorrelation function (ACF) \cite{ramsey1974characterization} are considered in hypothesis space. Despite widespread adoption, we find that these series-wise predictability metrics do not produce a significant correlation with final forecasting performance in terms of the Pearson coefficient, as shown in Figure \ref{fig:statistical-results}-left. This also indicates a long-standing gap between forecastability and forecasting methods, which can be partially closed by our proposed window-wise complexity.

\emph{(ii) Ablations on proposed window-wise complexity.} Here we also examine some variants of our complexity proposal, including replacing the frequency-domain formulation with time-domain alternatives such as window-wise Dynamic Time Warping (window DTW, \cite{berndt1994using}), as well as setting the time window size as past observation $P$ and forecasting horizon $F$. As illustrated in Figure \ref{fig:statistical-results}-middle, the frequency domain measure significantly outperforms the time-domain DTW, highlighting the merits of our definition in series pattern recognition. 
While using the forecasting window yields a higher Pearson coefficient, lower RESET test (Figure \ref{fig:statistical-results}-right) results suggest the presence of high-order dependencies, indicating that the relationship cannot be considered linear. Besides, we can also find that our design in considering both observation and prediction window is better than solely considering one of them, since considering length-$(P+F)$ series allows a comprehensive modeling of both margin distribution and covariance information.



\subsection{Practice 1: Identify Saturated Forecasting Tasks}

Forecasting benchmarks, such as ETT, electricity, and Weather \cite{zhou2021informer,wu2021autoformer}, have long served as benchmarks for deep forecasters. However, as presented in Figure~\ref{fig:model_performance}, recent models yield only marginal performance gains, suggesting that some widely used benchmarks may be approaching their intrinsic predictability limits. Identifying saturated tasks is crucial to enabling the community to shift focus from over-explored benchmarks and redirecting research toward more meaningful challenges. In this context, the proposed accuracy law provides an empirical lens for  this fundamental question.

\vspace{-10pt}
\paragraph{Practical usage} Building on extensive experiments, the accuracy law provides an empirical framework for estimating the achievable forecasting error of a given time series, enabling the identification of saturated benchmarks. By comparing the best performance of deep models on a benchmark with the error bound estimated by the accuracy law, we can assess whether the benchmark has reached its intrinsic predictability limit. When observed performance lies close to the fitted accuracy curve, the benchmark can be viewed as empirically approaching saturation.

\vspace{-10pt}
\paragraph{Results} Following the experiment protocol described in Section \ref{sec:exp_detail}, we compute the pattern complexity and best forecast performance of each variable within well-established benchmarks. As illustrated in Figure \ref{fig:saturation}, nearly all variables in ETT, Electricity, Weather, and Exchange-Rate benchmarks fall below the estimated performance bound, indicating that these datasets have likely reached their predictability limits. In contrast, a substantial fraction of variables in the Traffic benchmark remain above the fitted accuracy curve,which may be because this task is highly spatiotemporal dependent. This finding also implies a promising direction in spatiotemporal forecasting. 


\begin{figure*}[t]
    \centering
    \includegraphics[width=\linewidth]{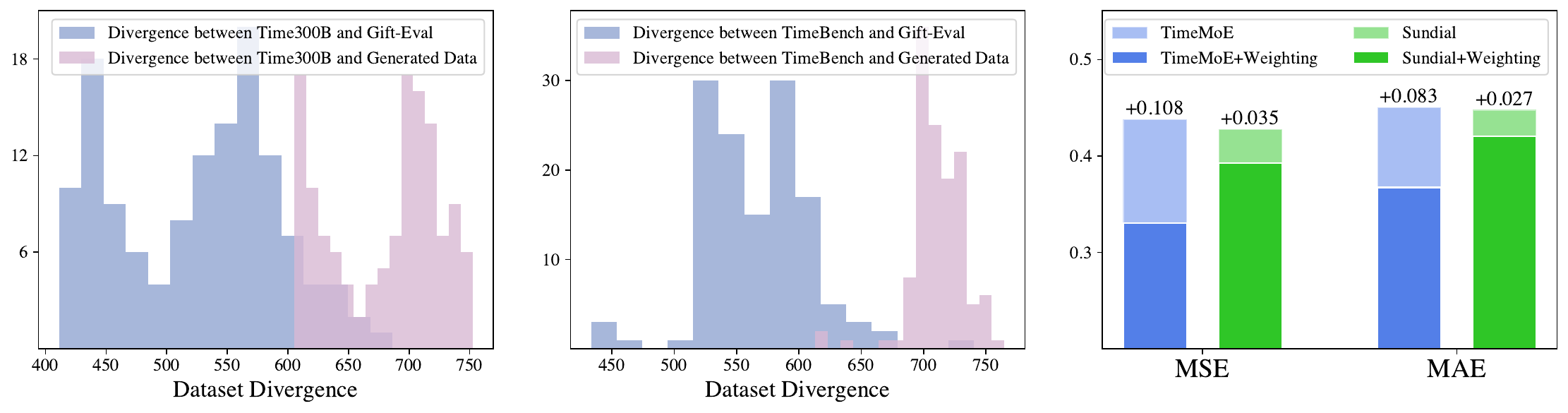}
    \vspace{-15pt}
    \caption{Practice of accuracy law in time series foundation models. Left: Histogram of divergence between pre-training corpus and GIFT-EVAL. Right: Performance comparison for weighted sampling strategy evaluated on our constructed generation data benchmark.}
    \label{fig:large-results}
    \vspace{-10pt}
\end{figure*}

\subsection{Practice 2: Guiding Time Series Foundation Models}


Despite their scale, existing pre-training corpora for time series foundation models remain highly imbalanced. For instance, Time-300B~\cite{shi2024time} is dominated by natural time series of moderate complexity, while high-complexity data such as web signals are severely underrepresented. This skew risks biasing models toward simpler patterns, thereby impairing generalization to complex dynamics.  However, current zero-shot benchmarks often struggle to reflect this limitation, as both pre-training and evaluation data are typically drawn from similar real-world sources, thereby obscuring out-of-distribution generalization.

\vspace{-5pt}
\paragraph{Usage 1: Constructing benchmark} In pursuit of a more challenging OOD evaluation benchmark, a key first step is to  quantify the pretraining-evaluation distribution gap. Surprisingly, we find that our proposed complexity metric (Eq.~\ref{equ:definition}) can be naturally extended to measure the divergence between two time series, further facilitating the evaluation of the domain gap. Specifically, given two time series $\mathbf{x}, \mathbf{y}$, the extension of Eq.~\ref{equ:definition} can be defined as:
\vspace{-5pt}
\begin{equation*}
\begin{aligned}
\operatorname{Divergence}(\mathbf{x}, \mathbf{y}) = \frac{1}{N M} \sum_{i=1}^N \sum_{j=1}^M \| \mathbf{A}_{i}^\mathbf{x} - \mathbf{A}_{j}^\mathbf{y} \|_2^2,\ \\
\text{where}\ \{\mathbf{A}_i^\ast\}_i = \left\{\operatorname{Amp}\big(\operatorname{FFT}\left(\mathbf{\ast}_{i:(i+P+F)}\right)\big)\right\}_i.
\end{aligned}
\end{equation*}
Based on the above metric, we quantify the divergence between the benchmark GIFT-Eval benchmark \cite{aksu2024gift}, and two advanced pretraining datasets, Time-300B~\cite{shi2024time} and TimeBench \cite{liu2025sundial}. Specifically, we randomly sample 100 series per subdomain from each dataset and compute the pairwise divergences. The resulting distributions are shown in Figure~\ref{fig:large-results}.

Guided by this metric,  we generate a new evaluation benchmark of 100,000 sequences following the procedure of \cite{ansari2024chronos}, explicitly reducing domain overlap with pre-training data. As demonstrated in Figure \ref{fig:large-results}-left, the resulting benchmark exhibits substantially higher complexity divergence than GIFT-Eval w.r.t.~the pretraining dataset, suggesting a potential stricter evaluation of the generalization capability of time series foundation models, such as Time-MoE \cite{shi2024time} and Sundial \cite{liu2025sundial}.


\vspace{-10pt}
\paragraph{Usage 2: Training strategy} Beyond constructing the evaluation benchmark, the complexity metric can also be used to derive a simple yet effective sampling strategy to ease the imbalance between quantity and pattern complexity of pre-training data. Specifically, by assigning higher sampling weights to time series with greater complexity calculated by Eq.~\ref{equ:definition}, the model is encouraged to learn more diverse and intricate temporal patterns during training.

To validate its effectiveness, we pre-train two representative time series foundation models, Time-MoE \cite{shi2024time} and Sundial \cite{liu2025sundial}, which feature point-wise and patch-wise modeling architectures, respectively, on the TimeBench dataset. We  evaluate their performance using our newly generated benchmark. To ensure a fair comparison, we strictly adhered to their reported pretraining paradigms, modifying only the data sampling weights. As illustrated in Figure \ref{fig:large-results}-right, the sampling strategy leads to consistent performance improvements across both models, demonstrating its generalizability and effectiveness.

\vspace{-5pt}
\section{Limitation}
In this paper, our empirical study is primarily confined to univariate time series forecasting. Therefore, the explored accuracy law does not yet account for scenarios where the forecasting performance relies heavily on exogenous factors or multivariate correlations. Future work will broaden this scope to incorporate additional performance metrics and extend to multivariate and multimodal forecasting settings.

\vspace{-5pt}
\section{Conclusion and Future Work}
In this paper, we empirically uncover an accuracy law for deep time series forecasters through thousands of controlled experiments and rigorous statistical analysis, revealing an exponential relationship between intrinsic window-wise complexity and the minimum achievable forecasting error of deep models. This empirical finding provides actionable insights for identifying saturated forecasting tasks and guiding the training of time series foundation models. To our knowledge, this is the first study to quantitatively characterize the relationship between time series complexity and forecasting performance in the context of deep learning. 

\section*{Impact Statement}
In this paper, we explore an accuracy law for deep time series forecasting. By introducing a window-wise complexity metric, we quantify the intrinsic difficulty of temporal patterns and uncover a consistent exponential relationship between this complexity and the minimum forecasting error of deep models. This finding offers critical guidance for the community, particularly in the development and evaluation of large-scale time series foundation models. While our work primarily focuses on advancing the theoretical and empirical understanding of Machine Learning, it has no obvious negative societal impacts.


\bibliography{example_paper}

@inproceedings{goerg2013forecastable,
  title={Forecastable component analysis},
  author={Goerg, Georg M},
  booktitle={Proceedings of the 30th International Conference on International Conference on Machine Learning},
  volume={28},
  pages={II--64},
  year={2013}
}

@article{wang2024deep,
  title={Deep time series models: A comprehensive survey and benchmark},
  author={Wang, Yuxuan and Wu, Haixu and Dong, Jiaxiang and Liu, Yong and Long, Mingsheng and Wang, Jianmin},
  journal={arXiv preprint arXiv:2407.13278},
  year={2024}
}

@article{ansari2024chronos,
  title={Chronos: Learning the language of time series},
  author={Ansari, Abdul Fatir and Stella, Lorenzo and Turkmen, Caner and Zhang, Xiyuan and Mercado, Pedro and Shen, Huibin and Shchur, Oleksandr and Rangapuram, Syama Sundar and Arango, Sebastian Pineda and Kapoor, Shubham and others},
  journal={arXiv preprint arXiv:2403.07815},
  year={2024}
}

@article{aksu2024gift,
  title={Gift-eval: A benchmark for general time series forecasting model evaluation},
  author={Aksu, Taha and Woo, Gerald and Liu, Juncheng and Liu, Xu and Liu, Chenghao and Savarese, Silvio and Xiong, Caiming and Sahoo, Doyen},
  journal={arXiv preprint arXiv:2410.10393},
  year={2024}
}

@article{nie2022time,
  title={A time series is worth 64 words: Long-term forecasting with transformers},
  author={Nie, Yuqi and Nguyen, Nam H and Sinthong, Phanwadee and Kalagnanam, Jayant},
  journal={arXiv preprint arXiv:2211.14730},
  year={2022}
}

@article{wang2024timemixer,
  title={Timemixer: Decomposable multiscale mixing for time series forecasting},
  author={Wang, Shiyu and Wu, Haixu and Shi, Xiaoming and Hu, Tengge and Luo, Huakun and Ma, Lintao and Zhang, James Y and Zhou, Jun},
  journal={arXiv preprint arXiv:2405.14616},
  year={2024}
}

@inproceedings{zeng2023transformers,
  title={Are transformers effective for time series forecasting?},
  author={Zeng, Ailing and Chen, Muxi and Zhang, Lei and Xu, Qiang},
  booktitle={Proceedings of the AAAI conference on artificial intelligence},
  volume={37},
  pages={11121--11128},
  year={2023}
}

@inproceedings{gao2023stockformer,
  title={StockFormer: learning hybrid trading machines with predictive coding},
  author={Gao, Siyu and Wang, Yunbo and Yang, Xiaokang},
  booktitle={International Joint Conference on Artificial Intelligence},
  pages={4766--4774},
  year={2023}
}

@article{wang2024news,
  title={From news to forecast: Integrating event analysis in llm-based time series forecasting with reflection},
  author={Wang, Xinlei and Feng, Maike and Qiu, Jing and Gu, Jinjin and Zhao, Junhua},
  journal={Advances in Neural Information Processing Systems},
  volume={37},
  pages={58118--58153},
  year={2024}
}

@inproceedings{guo2019attention,
  title={Attention based spatial-temporal graph convolutional networks for traffic flow forecasting},
  author={Guo, Shengnan and Lin, Youfang and Feng, Ning and Song, Chao and Wan, Huaiyu},
  booktitle={Proceedings of the AAAI conference on artificial intelligence},
  volume={33},
  pages={922--929},
  year={2019}
}

@inproceedings{wu2023timesnet,
  title={TimesNet: Temporal 2D-Variation Modeling for General Time Series Analysis},
  author={Haixu Wu and Tengge Hu and Yong Liu and Hang Zhou and Jianmin Wang and Mingsheng Long},
  booktitle={International Conference on Learning Representations},
  year={2023},
}

@incollection{benesty2009pearson,
  title={Pearson correlation coefficient},
  author={Benesty, Jacob and Chen, Jingdong and Huang, Yiteng and Cohen, Israel},
  booktitle={Noise reduction in speech processing},
  year={2009},
  publisher={Springer}
}

@article{yu2017spatio,
  title={Spatio-temporal graph convolutional networks: A deep learning framework for traffic forecasting},
  author={Yu, Bing and Yin, Haoteng and Zhu, Zhanxing},
  journal={arXiv preprint arXiv:1709.04875},
  year={2017}
}

@article{wu2023interpretable,
  title={Interpretable weather forecasting for worldwide stations with a unified deep model},
  author={Wu, Haixu and Zhou, Hang and Long, Mingsheng and Wang, Jianmin},
  journal={Nature Machine Intelligence},
  volume={5},
  number={6},
  pages={602--611},
  year={2023},
  publisher={Nature Publishing Group UK London}
}

@inproceedings{woo2024unified,
  title={Unified Training of Universal Time Series Forecasting Transformers},
  author={Woo, Gerald and Liu, Chenghao and Kumar, Akshat and Xiong, Caiming and Savarese, Silvio and Sahoo, Doyen},
  booktitle={International Conference on Machine Learning},
  year={2024}
}

@article{ramsey1974characterization,
  title={Characterization of the partial autocorrelation function},
  author={Ramsey, Fred L},
  journal={The Annals of Statistics},
  pages={1296--1301},
  year={1974},
  publisher={JSTOR}
}

@article{volkova2013research,
  title={The research of distribution of the Ramsey RESET-test statistic},
  author={Volkova, VM and Pankina, VL},
  year={2013}
}

@inproceedings{berndt1994using,
  title={Using dynamic time warping to find patterns in time series},
  author={Berndt, Donald J and Clifford, James},
  booktitle={International conference on knowledge discovery and data mining},
  pages={359--370},
  year={1994}
}

@inproceedings{deng2009imagenet,
  title={Imagenet: A large-scale hierarchical image database},
  author={Deng, Jia and Dong, Wei and Socher, Richard and Li, Li-Jia and Li, Kai and Fei-Fei, Li},
  booktitle={2009 IEEE conference on computer vision and pattern recognition},
  pages={248--255},
  year={2009},
  organization={Ieee}
}

@article{krizhevsky2012imagenet,
  title={Imagenet classification with deep convolutional neural networks},
  author={Krizhevsky, Alex and Sutskever, Ilya and Hinton, Geoffrey E},
  journal={Advances in neural information processing systems},
  volume={25},
  year={2012}
}

@article{simonyan2014very,
  title={Very deep convolutional networks for large-scale image recognition},
  author={Simonyan, Karen and Zisserman, Andrew},
  journal={arXiv preprint arXiv:1409.1556},
  year={2014}
}

@inproceedings{he2016deep,
  title={Deep residual learning for image recognition},
  author={He, Kaiming and Zhang, Xiangyu and Ren, Shaoqing and Sun, Jian},
  booktitle={Proceedings of the IEEE conference on computer vision and pattern recognition},
  pages={770--778},
  year={2016}
}

@article{bi2023accurate,
  title={Accurate medium-range global weather forecasting with 3D neural networks},
  author={Bi, Kaifeng and Xie, Lingxi and Zhang, Hengheng and Chen, Xin and Gu, Xiaotao and Tian, Qi},
  journal={Nature},
  volume={619},
  number={7970},
  pages={533--538},
  year={2023},
  publisher={Nature Publishing Group UK London}
}

@article{yang2020qlib,
  title={Qlib: An ai-oriented quantitative investment platform},
  author={Yang, Xiao and Liu, Weiqing and Zhou, Dong and Bian, Jiang and Liu, Tie-Yan},
  journal={arXiv preprint arXiv:2009.11189},
  year={2020}
}

@book{cramer1999mathematical,
  title={Mathematical methods of statistics},
  author={Cram{\'e}r, Harald},
  volume={9},
  year={1999},
  publisher={Princeton university press}
}

@book{kay1993fundamentals,
  title={Fundamentals of statistical signal processing: estimation theory},
  author={Kay, Steven M},
  year={1993},
  publisher={Prentice-Hall, Inc.}
}

@misc{elliott1992efficient,
  title={Efficient tests for an autoregressive unit root},
  author={Elliott, Graham and Rothenberg, Thomas J and Stock, James H},
  year={1992},
  publisher={National Bureau of Economic Research Cambridge, Mass., USA}
}

@article{pincus1991approximate,
  title={Approximate entropy as a measure of system complexity.},
  author={Pincus, Steven M},
  journal={Proceedings of the national academy of sciences},
  volume={88},
  number={6},
  pages={2297--2301},
  year={1991}
}

@article{richman2000physiological,
  title={Physiological time-series analysis using approximate entropy and sample entropy},
  author={Richman, Joshua S and Moorman, J Randall},
  journal={American journal of physiology-heart and circulatory physiology},
  volume={278},
  number={6},
  pages={H2039--H2049},
  year={2000},
  publisher={American Physiological Society Bethesda, MD}
}

@inproceedings{zhou2021informer,
  title={Informer: Beyond efficient transformer for long sequence time-series forecasting},
  author={Zhou, Haoyi and Zhang, Shanghang and Peng, Jieqi and Zhang, Shuai and Li, Jianxin and Xiong, Hui and Zhang, Wancai},
  booktitle={Proceedings of the AAAI conference on artificial intelligence},
  volume={35},
  pages={11106--11115},
  year={2021}
}

@article{wu2021autoformer,
  title={Autoformer: Decomposition transformers with auto-correlation for long-term series forecasting},
  author={Wu, Haixu and Xu, Jiehui and Wang, Jianmin and Long, Mingsheng},
  journal={Advances in neural information processing systems},
  volume={34},
  pages={22419--22430},
  year={2021}
}

@inproceedings{lai2018modeling,
  title={Modeling long-and short-term temporal patterns with deep neural networks},
  author={Lai, Guokun and Chang, Wei-Cheng and Yang, Yiming and Liu, Hanxiao},
  booktitle={The 41st international ACM SIGIR conference on research \& development in information retrieval},
  pages={95--104},
  year={2018}
}

@article{liu2022scinet,
  title={Scinet: Time series modeling and forecasting with sample convolution and interaction},
  author={Liu, Minhao and Zeng, Ailing and Chen, Muxi and Xu, Zhijian and Lai, Qiuxia and Ma, Lingna and Xu, Qiang},
  journal={Advances in Neural Information Processing Systems},
  volume={35},
  pages={5816--5828},
  year={2022}
}

@book{box2015time,
  title={Time series analysis: forecasting and control},
  author={Box, George EP and Jenkins, Gwilym M and Reinsel, Gregory C and Ljung, Greta M},
  year={2015},
  publisher={John Wiley \& Sons}
}

@book{hamilton2020time,
  title={Time series analysis},
  author={Hamilton, James D},
  year={2020},
  publisher={Princeton university press}
}

@article{granger2001spurious,
  title={Spurious regressions with stationary series},
  author={Granger IV, Clive WJ and Hyung, Namwon and Jeon, Yongil},
  journal={Applied Economics},
  year={2001},
  publisher={Taylor \& Francis}
}

@article{liu2025sundial,
  title={Sundial: A family of highly capable time series foundation models},
  author={Liu, Yong and Qin, Guo and Shi, Zhiyuan and Chen, Zhi and Yang, Caiyin and Huang, Xiangdong and Wang, Jianmin and Long, Mingsheng},
  journal={International Conference on Machine Learning},
  year={2025}
}

@article{shi2024time,
  title={Time-moe: Billion-scale time series foundation models with mixture of experts},
  author={Shi, Xiaoming and Wang, Shiyu and Nie, Yuqi and Li, Dianqi and Ye, Zhou and Wen, Qingsong and Jin, Ming},
  journal={arXiv preprint arXiv:2409.16040},
  year={2024}
}

@inproceedings{stitsyuk2025xpatch,
  title={xPatch: Dual-Stream Time Series Forecasting with Exponential Seasonal-Trend Decomposition},
  author={Stitsyuk, Artyom and Choi, Jaesik},
  booktitle={Proceedings of the AAAI Conference on Artificial Intelligence},
  volume={39},
  pages={20601--20609},
  year={2025}
}

@article{wang2024timemixer++,
  title={Timemixer++: A general time series pattern machine for universal predictive analysis},
  author={Wang, Shiyu and Li, Jiawei and Shi, Xiaoming and Ye, Zhou and Mo, Baichuan and Lin, Wenze and Ju, Shengtong and Chu, Zhixuan and Jin, Ming},
  journal={arXiv preprint arXiv:2410.16032},
  year={2024}
}

@article{kaplan2020scaling,
  title={Scaling laws for neural language models},
  author={Kaplan, Jared and McCandlish, Sam and Henighan, Tom and Brown, Tom B and Chess, Benjamin and Child, Rewon and Gray, Scott and Radford, Alec and Wu, Jeffrey and Amodei, Dario},
  journal={arXiv preprint arXiv:2001.08361},
  year={2020}
}

@article{hoffmann2022training,
  title={Training compute-optimal large language models},
  author={Hoffmann, Jordan and Borgeaud, Sebastian and Mensch, Arthur and Buchatskaya, Elena and Cai, Trevor and Rutherford, Eliza and Casas, Diego de Las and Hendricks, Lisa Anne and Welbl, Johannes and Clark, Aidan and others},
  journal={arXiv preprint arXiv:2203.15556},
  year={2022}
}
\bibliographystyle{icml2026}

\newpage
\appendix
\onecolumn
\section{Supplement Results for Figure \ref{fig:accuracy-law}}
\begin{figure}[h]
    \centering

    \begin{subfigure}[b]{\linewidth}
        \centering
        \includegraphics[width=\linewidth]{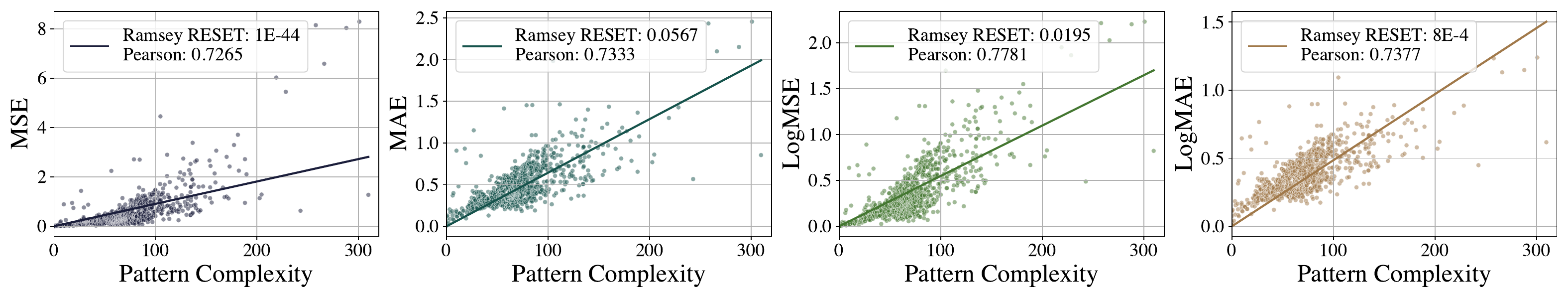} 
        \subcaption{Experimental results using DLinear. The fitted coefficient $\alpha=0.0055$.}
        \label{subfig:model2}
    \end{subfigure}
    
    \begin{subfigure}[b]{\linewidth}
        \centering
        \includegraphics[width=\linewidth]{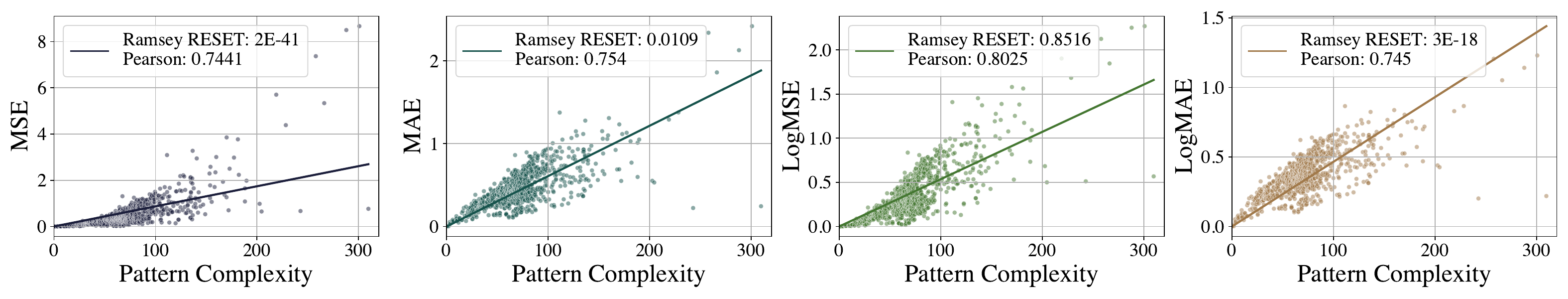}  
        \subcaption{Experimental results using DLinear and PatchTST. The fitted coefficient $\alpha=0.0054$.}
        \label{subfig:model3}
    \end{subfigure}

    \begin{subfigure}[b]{\linewidth}
        \centering
        \includegraphics[width=\linewidth]{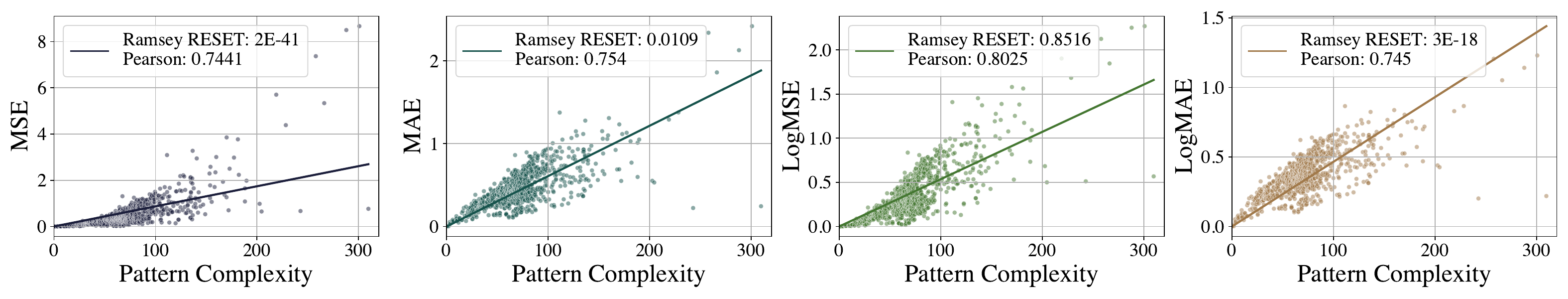}
        \subcaption{Experimental results using DLinear, PatchTST and TimeMixer. The fitted coefficient $\alpha=0.0054$.}
        \label{subfig:model4}
    \end{subfigure}

    \begin{subfigure}[b]{\linewidth}
        \centering
        \includegraphics[width=\linewidth]{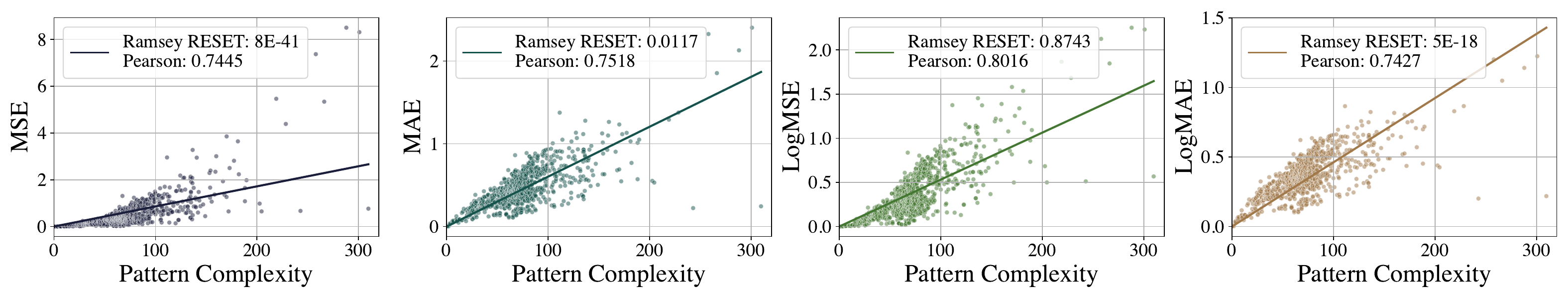} 
        \subcaption{Experimental results using DLinear, PatchTST, TimeMixer, and xPatch. The fitted coefficient $\alpha=0.0054$. }
        \label{subfig:model5}
    \end{subfigure}
    
    \caption{Experimental performance and statistical test based on different number of deep time series models, which is in the vertical layout.}
    \label{fig:acclaw_increase_model}
\end{figure}

In this section, we validate the generalizability of the proposed accuracy law and ensure it is not a special case biased by specific model selections.
Concretely, we simulate the evolution of deep forecasters by establishing a fixed, chronological order for the constituent models used to fit Figure \ref{fig:accuracy-law}: DLinear, PatchTST, TimeMixer, xPatch, and TimeMixerPP. Starting with DLinear, we cumulatively add the new models to fit the accuracy curve. Figure \ref{fig:acclaw_increase_model} demonstrates that the introduction of advanced architectures does not alter the fundamental exponential relationship between pattern complexity and forecasting performance. The evolution of models merely results in marginal variations in the coefficient $\alpha$, confirming the robustness of our empirical findings.

\section{Supplement Results for Figure \ref{fig:statistical-results}}

\subsection{Experimental Results for Classical Predictability Metrics}
\begin{figure}[h]
    \centering
    \includegraphics[width=\linewidth]{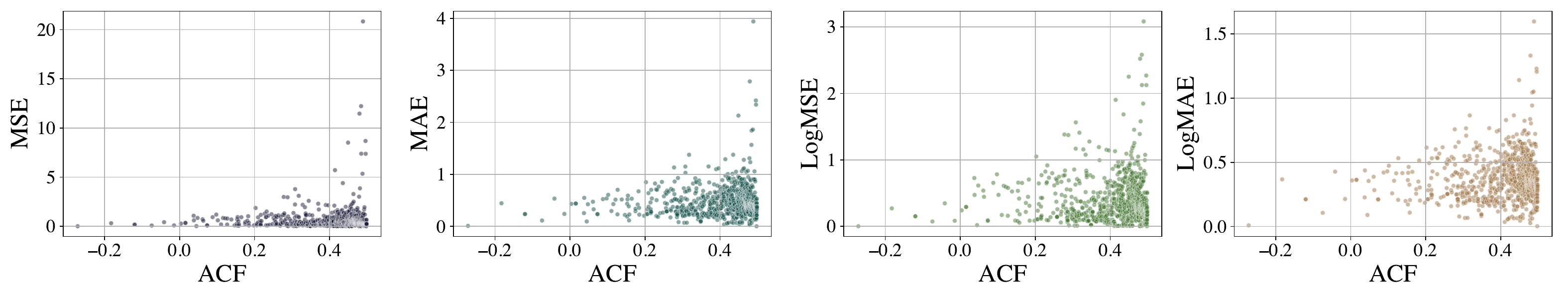}
    \includegraphics[width=\linewidth]{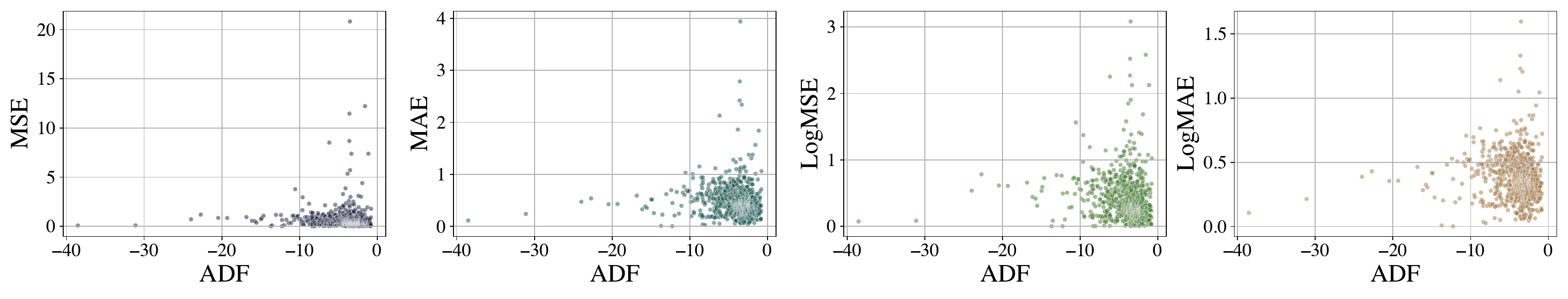}
    \includegraphics[width=\linewidth]{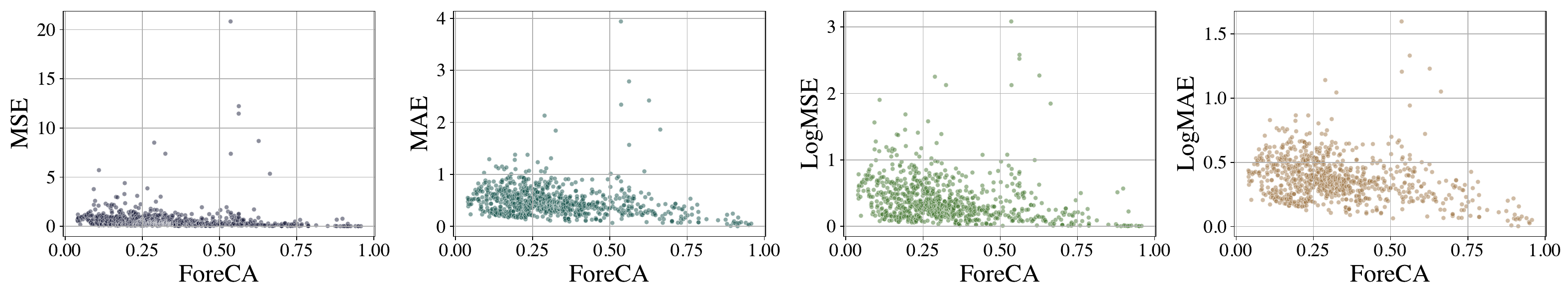}
    \vspace{-15pt}
    \caption{Experimental results between minimum forecasting error and classical forecasting metrics.}
    \vspace{-5pt}
    \label{fig:app-vis-metric}
\end{figure}

We visualize the experimental results from 940 series and 2,820 deep forecasters for three classical predictability metrics. As shown in Figure \ref{fig:app-vis-metric}, the distribution of the points appears highly disordered, showing no clear correlation between these traditional metrics and the optimal forecasting error, even when considering more complex relations beyond the exponential and linear assumptions within our hypothesis space. This further validates the challenges and efforts of deriving the accuracy law.

\subsection{Quantitative Results}

We perform extensive statistical analysis to examine the linear relationship between forecasting performance and different forecasting metrics, using the Pearson correlation coefficient \citep{benesty2009pearson} and the Ramsey RESET test \citep{volkova2013research}. The detailed results are presented in Table \ref{tab:statistics}. A higher Pearson correlation coefficient indicates a stronger positive linear relationship. Beyond the linear dependence analysis, the Ramsey RESET test provides a more comprehensive evaluation of whether higher-order dependencies exist. Statistically, a p-value of 0.05 serves as the critical threshold for significance. When the p-value of the Ramsey RESET test is larger than 0.05, the underlying relationship can be described as a pure linear relation.

\begin{table}[htbp]
    \small
    \centering
    \caption{Quantitative results of statistical test with different metrics. For clarity, we mark the value with \colorbox{lightgray}{gray} if its Pearson Correlation coefficient is below 0.6 or the Ramsey RESET p-value falls below the 0.05 significance threshold, which means there does not exist a pure linear relation.}
    \vspace{-5pt}
    \setlength{\tabcolsep}{4pt}
    \begin{tabular*}{\textwidth}{@{\extracolsep{\fill}}lcccccccc@{}}
        \toprule
        \multirow{2}{*}{Metric} & \multicolumn{4}{c}{Pearson Correlation} & \multicolumn{4}{c}{Ramsey RESET} \\
        \cmidrule(lr){2-5} \cmidrule(lr){6-9}
        & MSE & MAE & LogMSE & LogMAE & MSE & MAE & LogMSE & LogMAE \\
        \midrule
        ACF & \cellcolor{lightgray}{0.0444} & \cellcolor{lightgray}{0.1040} & \cellcolor{lightgray}{0.0309} & \cellcolor{lightgray}{0.1115} & 9.78E-01	& 4.38E-01	 & 1.53E-01 & 1.92E-01 \\
        ADF & \cellcolor{lightgray}{-0.0428} & \cellcolor{lightgray}{-0.0436} & \cellcolor{lightgray}{-0.1165} & \cellcolor{lightgray}{-0.0487} & \cellcolor{lightgray}{9.23E-02} & \cellcolor{lightgray}{4.51E-04} & \cellcolor{lightgray}{9.28E-05} & \cellcolor{lightgray}{5.00E-05} \\
        ForeCA & \cellcolor{lightgray}{-0.0541} & \cellcolor{lightgray}{-0.211} & \cellcolor{lightgray}{-0.2283} & \cellcolor{lightgray}{-0.292} & 8.53E-01 & \cellcolor{lightgray}{1.07E-02} & \cellcolor{lightgray}{3.17E-01} & \cellcolor{lightgray}{3.72E-04} \\
        \midrule
        DTW & \cellcolor{lightgray}{0.545} & \textbf{0.6055} & \cellcolor{lightgray}{0.5786} & \cellcolor{lightgray}{0.5733} & \cellcolor{lightgray}{1.36E-31} & \textbf{7.34E-01} & 8.82E-02 & \cellcolor{lightgray}{2.03E-05} \\
        Ours (Past Only) & \cellcolor{lightgray}{0.4492} & \cellcolor{lightgray}{0.5875} & \cellcolor{lightgray}{0.5416} & \cellcolor{lightgray}{0.5848} & \cellcolor{lightgray}{1.75E-19} & \cellcolor{lightgray}{4.95E-02} & \cellcolor{lightgray}{6.87E-05} & 5.50E-02 \\
        Ours (Forecast Only) & 0.7751 & 0.7639 & 0.8079 & 0.7520 & \cellcolor{lightgray}{6.95E-54} & \cellcolor{lightgray}{2.20E-02} & 5.56E-02 & \cellcolor{lightgray}{3.34E-15} \\
        Ours (Past+Forecast) & 0.7441 & 0.7540 & \textcolor{black}{\textbf{0.8025}} & 0.7450 & \cellcolor{lightgray}{2.16E-41} & \cellcolor{lightgray}{1.09E-02} & \textcolor{black}{\textbf{8.52E-01}} & \cellcolor{lightgray}{3.16E-18} \\
        \bottomrule
    \end{tabular*}
    \label{tab:statistics}
\end{table}

\section{Supplement Results for Figure \ref{fig:saturation}}
In this section, we present the detailed saturation analysis on existing long-term forecasting benchmarks. The results listed in Table \ref{tab:staturation} show that 100\% of the variables in the ETTh2, ETTm2, Weather and Exchange-Rate benchmarks fall below the estimated performance bound, suggesting that these datasets can be considered saturated. In contrast, the saturation variables in the Traffic dataset remain below 80\%, suggesting that this dataset still holds potential for further exploration and improvement. 

\begin{table}[htbp]
    \small
  \centering
  \caption{Saturation test results on existing long-term forecasting benchmarks, where we count the number of variables that fall below the performance estimated by the accuracy law.}
  \vspace{-5pt}
  \setlength{\tabcolsep}{2pt}
  \begin{tabular*}{\textwidth}{@{\extracolsep{\fill}}lcccccccc@{}}
    \toprule  
    Dataset       & ETTh1    & ETTh2    & ETTm1     & ETTm2     & Electricity & Traffic   & Weather   & Exchange Rate \\
    \midrule 
    Total          & 7        & 7        & 7         & 7         & 321         & 862       & 21        & 8             \\
    Statured Count & 4        & 7        & 6         & 7         & 314         & 677       & 20        & 8             \\
    Statured Ratio & 57.14\%  & 100\%    & 85.71\%   & 100\%     & 97.82\%     & 78.54\%   & 85.24\%     & 100\%         \\
    \bottomrule  
  \end{tabular*}
  \label{tab:staturation}
\end{table}

\begin{figure}[h]
    \centering
    \includegraphics[width=\linewidth]{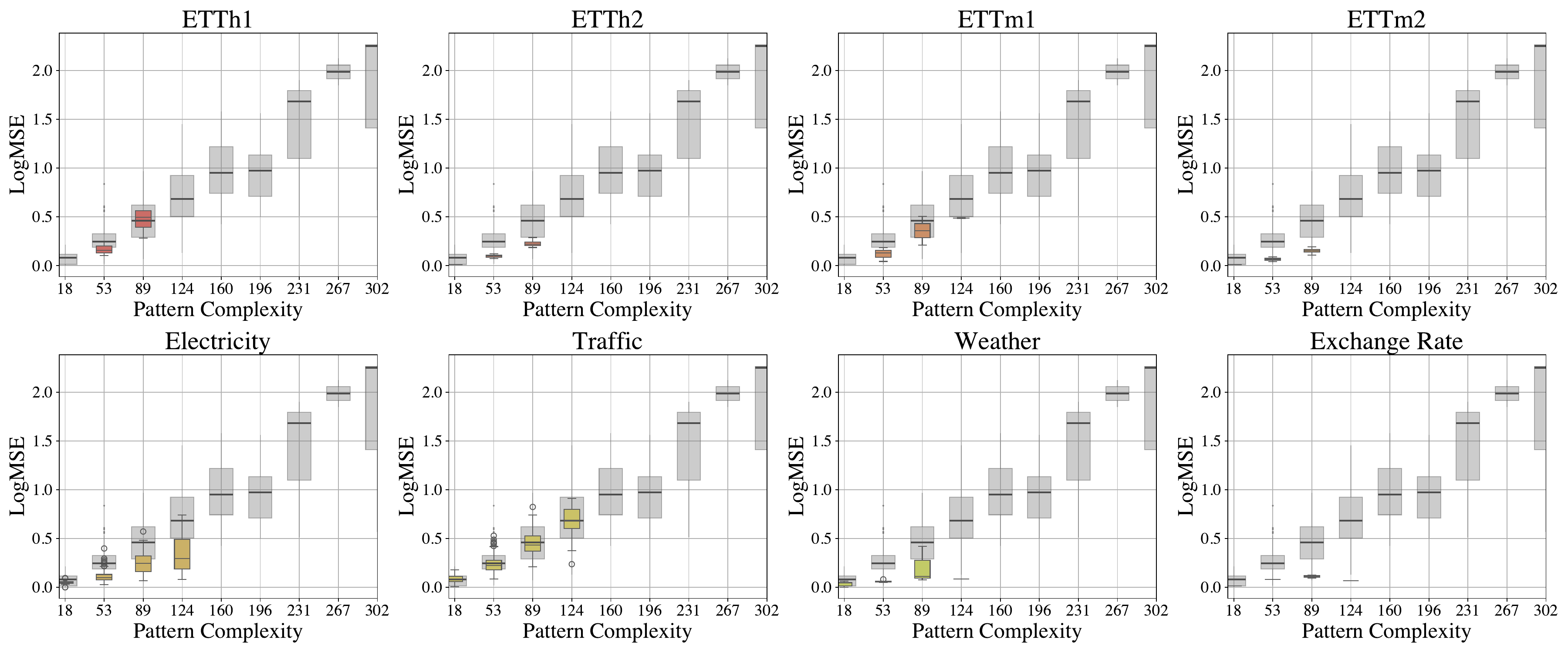}
    \caption{Saturation analysis under boxed visualization. Gray boxes refer to the data points from LOTSA \citeyearpar{woo2024unified} used in computing the accuracy law, while others refer to different benchmarks.}
    \label{fig:app-box-vis}
    \vspace{-10pt}
\end{figure}

Additionally, we provide a box plot visualization comparing each benchmark with the data used in testing the accuracy law, as shown in Figure \ref{fig:app-box-vis}. Notably, all variables in the Exchange-Rate dataset lie entirely below the lower quartile of the reference box plot distribution, underscoring its significant deviation from expected performance levels.

\section{Implementation Details}

\subsection{Classical Predictability Metrics}
In our hypothesis space, we consider three conventional time series predictability metrics.

\vspace{-5pt}
\paragraph{ADF} The augmented Dickey–Fuller test (ADF) \citep{elliott1992efficient} is a widely-used statistical test to assess the predictability of time series, as stationarity is a fundamental property of time series that significantly influences its predictability. For a time series $\mathbf{x}$, the ADF test operates under the null hypothesis that there is a unit
root presents in time series using the following regression model:
\begin{equation}
    \begin{aligned}
        \Delta \mathbf{x}_t = \alpha + \beta t + \gamma \mathbf{x}_{t-1} + \sum_{i=1}^p \phi_i \Delta \mathbf{x}_{t-i} + \epsilon_t
    \end{aligned}
\end{equation}
where $\alpha$ is a constant, $\beta$ is the coefficient on a time trend, and $\gamma$ is the coefficient testing the presence of a unit root. A significantly negative test statistic for $\gamma$ indicates rejection of the null hypothesis, suggesting stationarity. Time series deemed stationary by the ADF test are generally more predictable due to stable statistical properties over time.

\vspace{-5pt}
\paragraph{ACF} The autocorrelation function (ACF) \citep{goerg2013forecastable} is a classical tool for measuring the degree of correlation between a time series and its lagged values over time, which is critical for determining its predictability.
Given a time series $\mathbf{x}$, the autocorrelation with lag-$\tau$ is defined as:
\begin{equation}
\operatorname{ACF}(\tau) = \frac{\operatorname{Cov}(\mathbf{x}_t, \mathbf{x}_{t-\tau})}{\sqrt{\operatorname{Var}(\mathbf{x}_t)\operatorname{Var(\mathbf{x}_{t-\tau})}}}.
\end{equation}
Here, we adopt the half-life length of the autocorrelation function defined above, which is the value of $\tau$, such that $\operatorname{ACF}(\tau)=\frac{1}{2}\operatorname{ACF}(1)$. This metric is widely used in financial analysis.

\vspace{-5pt}
\paragraph{ForeCA} Given a time series $\mathbf{x}$, the forecastability \citep{goerg2013forecastable} can be quantified by analyzing the degree of orderliness in the frequency domain. Specifically, it is computed by subtracting the entropy of the normalized amplitudes of the series in the frequency domain as follows:
\begin{equation}
    \begin{aligned}
        \textbf{A} &= \operatorname{Amp}\big(\operatorname{FFT}\left(\mathbf{x}\right)\big), \mathbf{p}_k = \frac{\mathbf{A}_k}{\sum_{k=1}^L(\mathbf{A}_k)}, \operatorname{ForeCA}(\mathbf{x})= 1 - \operatorname{Entropy}(\mathbf{p}),
    \end{aligned}
\end{equation}
where $\operatorname{Amp}(\cdot)$ represents the amplitude spectrum and $\mathbf{A}_k$ denotes the intensity of the frequency-$k$ periodic basis function. The forecastability metric $\operatorname{ForeCA}$ is derived by subtracting the entropy from 1, with a value closer to 1 indicating a highly predictable time series, and a value closer to 0 indicating a highly chaotic or random series that is less predictable.

\subsection{Statistical Test}
This section presents details of the statistical tests employed.

\vspace{-5pt}
\paragraph{Pearson correlation}
The Pearson correlation coefficient \citep{benesty2009pearson} is widely used to quantify the strength and direction of the \emph{linear relationship} between two variables, $\mathbf{x}$ and $\mathbf{y}$. It is calculated as the covariance of the two variables divided by the product of their standard deviations:
\begin{equation}
    \operatorname{Pearson}(\mathbf{x}, \mathbf{y}) = \frac{\sum_{i=1}^N \left( \mathbf{x}_i - \bar{\mathbf{x}} \right) \left( \mathbf{y}_i - \bar{\mathbf{y}} \right)}{\sqrt{\sum_{i=1}^N \left( \mathbf{x}_i - \bar{\mathbf{x}} \right)^2 \sum_{i=1}^N \left( \mathbf{y}_i - \bar{\mathbf{y}} \right)^2}},
\end{equation}
where $\bar{\mathbf{x}} = \frac{1}{N} \sum_{i=1}^N \mathbf{x}_i,\, \bar{\mathbf{y}} = \frac{1}{N} \sum_{i=1}^N \mathbf{y}_i$ are the sample means. The coefficient ranges from -1 (perfect negative linear relationship) to +1 (perfect positive linear relationship). 

\vspace{-5pt}
\paragraph{Ramsey RESET test} The Ramsey RESET test \citep{volkova2013research} is frequently used to detect model specification errors, such as omitted higher-order terms in a regression model. In our statistical tests, we use it to test whether the second-order term is statistically required to precisely describe the relationship. Concretely, the procedure of Ramsey RESET for the following linear regression model is shown here, 
\[
y = \beta_0 + \beta_1 x_1 + \beta_2 x_2 + \dots + \beta_k x_k + \epsilon.
\]
\emph{Step 1:} Fit the linear regression model and compute the predicted value $\hat{y}$ from the model. Then calculate the \(R^2_{\text{old}}\) from the linear model.

\emph{Step 2:} Perform a regression with an extended model that includes the high-order predicted values and calculate the new \(R^2_{\text{new}}\).
\[
y = \beta_0 + \beta_1 x_1 + \beta_2 x_2 + \dots + \beta_k x_k + \gamma_1 \hat{y}^2 + \gamma_2 \hat{y}^3 + \dots + \gamma_m \hat{y}^m + \nu.
\]
\emph{Step 3:} The Ramsey RESET test statistic can be calculated through,
\begin{equation}
\operatorname{Ramsey\ RESET} = \frac{(R^2_{\text{new}} - R^2_{\text{old}}) / m}{(1 - R^2_{\text{new}}) / (n - k - m -1)},
\end{equation}
where \(n\) denotes the sample size, \(k\) refer to the dimension in the linear model, and \(m\) is the number of higher-order terms added to the extended model.

Theorectically, the statistic \(\operatorname{Ramsey\ RESET}\) follows an \(F\)-distribution with \(m\) and \((n - k - m-1)\) degrees of freedom. To obtain the p-value, we compare the observed statistic to the standard F-distribution. If the p-value is less than the significance level (e.g., 0.05), we reject the null hypothesis and conclude that the original model may suffer from specification errors, such as missing important nonlinear terms or interaction effects. Conversely, there is insufficient evidence to suggest specification errors in the original model based on this test.

\section{Analysis of conventional series-wise indicators} 

\begin{figure*}[h]
    \centering
    \includegraphics[width=\linewidth]{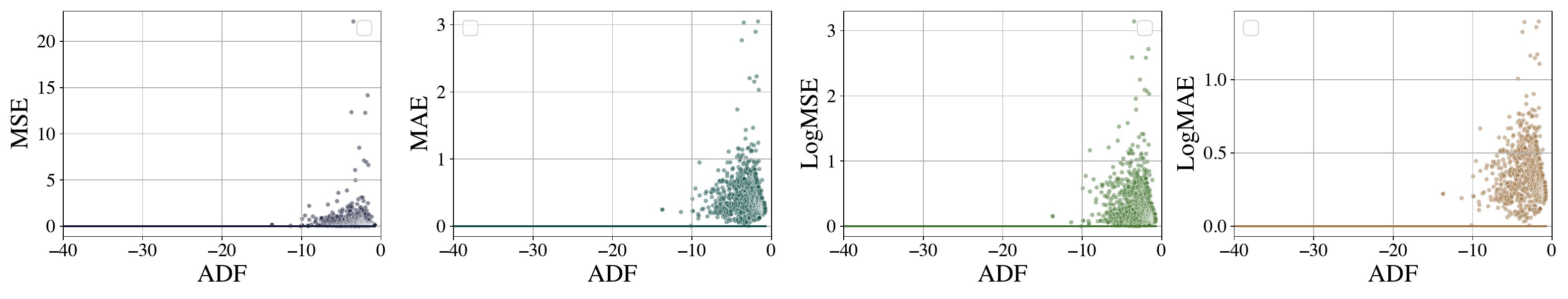}
    \caption{Relation between series-wise ADF and model performance under increased input length.}
    \label{fig:adf_512}
    \vspace{-5pt}
\end{figure*}

As discussed before, the classical series-wise indicators do not match the sequence-to-sequence deep forecasting paradigm. Here, to enhance the classical indicators, we increase the input length of deep models to 512 and explore whether there is some correlation between ADF and performance. The result is in Figure \ref{fig:adf_512}, where still no obvious relation can be observed, highlighting the importance of window-wise complexity definition.

\section{Individual Performance of Five Experimental Models}
In the experiments, the forecasting performance for each series was defined as the minimum forecasting errors across several advanced deep forecasters. Here we further explore the relationship between series pattern complexity and the forecasting performance of each individual model. We can find that the relation between complexity and performance persists. The Pearson correlation coefficients for the individual models are 0.7979, 0.7772, 0.7647, 0.7236, and 0.7915 respectively.

\begin{figure}[h]
    \centering
    \begin{subfigure}[b]{\linewidth} 
        \centering
        \includegraphics[width=\linewidth]{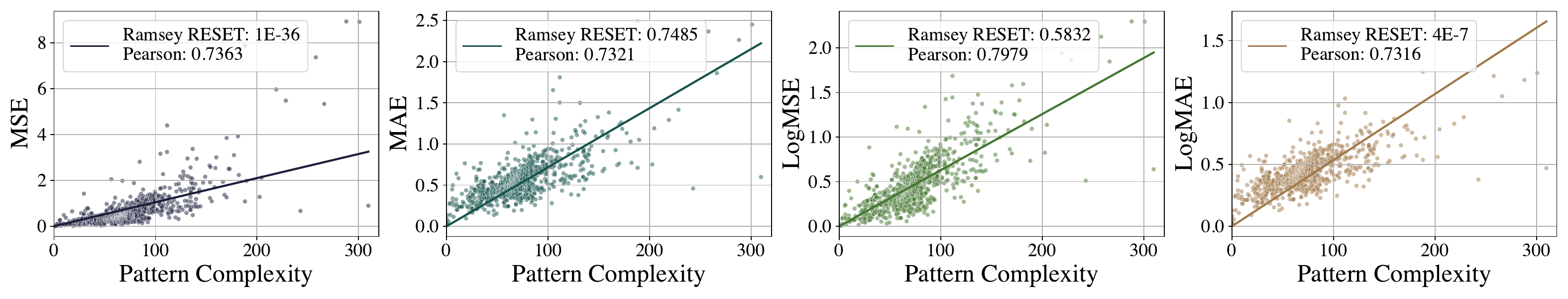} 
        \subcaption{DLinear}
    \end{subfigure}
    
    \begin{subfigure}[b]{\linewidth}
        \centering
        \includegraphics[width=\linewidth]{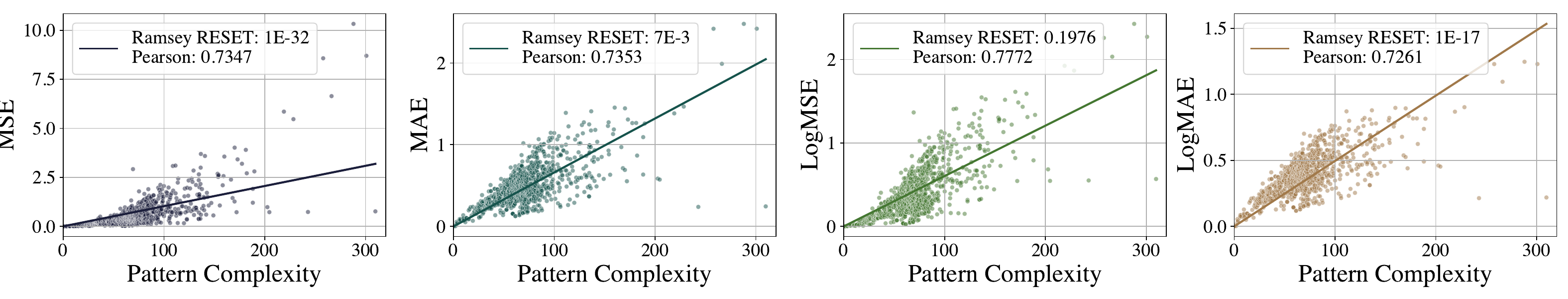} 
        \subcaption{PatchTST}
    \end{subfigure}
    
    \begin{subfigure}[b]{\linewidth}
        \centering
        \includegraphics[width=\linewidth]{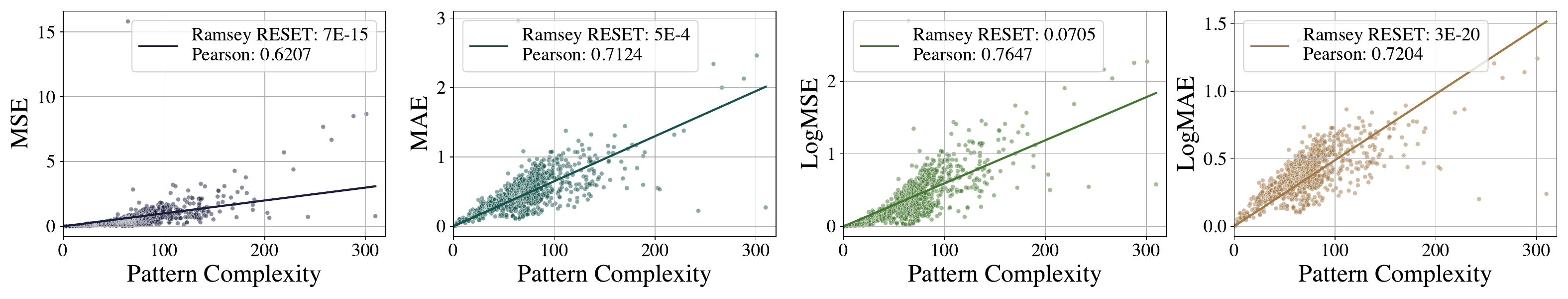}  
        \subcaption{TimeMixer}
    \end{subfigure}

    \begin{subfigure}[b]{\linewidth}
        \centering
        \includegraphics[width=\linewidth]{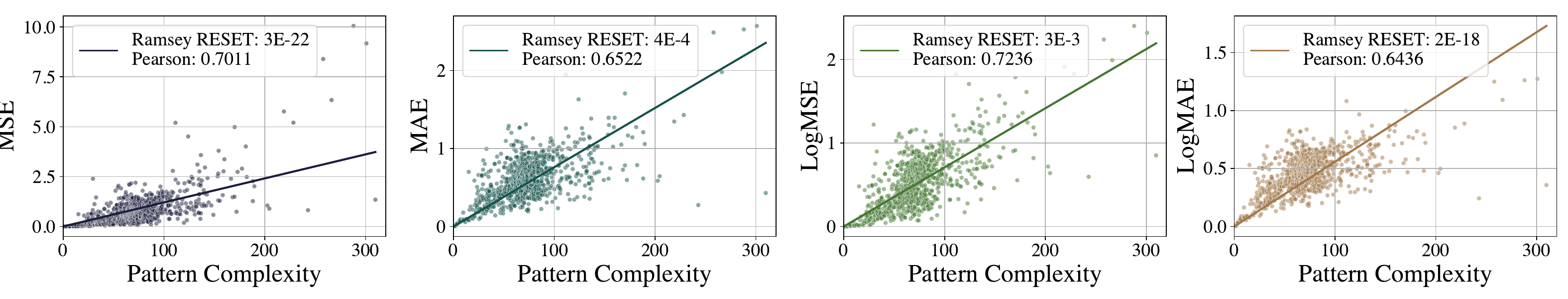}
        \subcaption{TimeMixer++}
    \end{subfigure}

    \begin{subfigure}[b]{\linewidth}
        \centering
        \includegraphics[width=\linewidth]{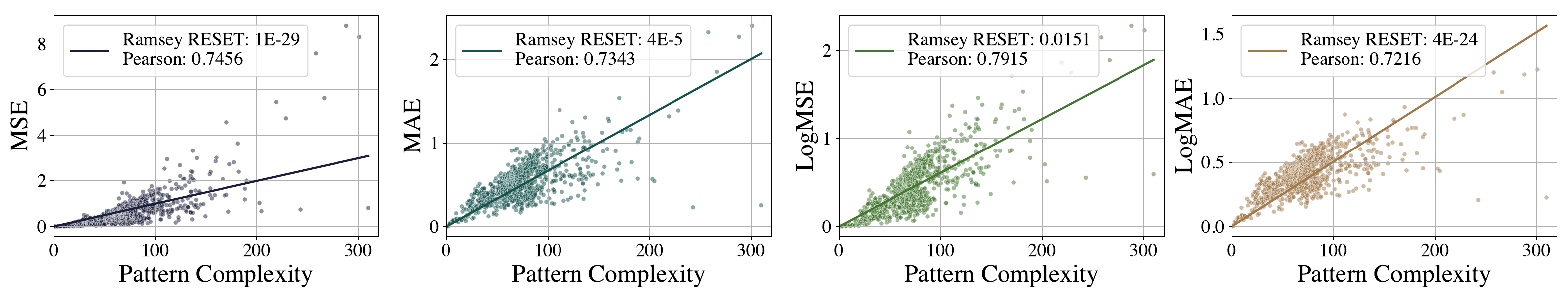} 
        \subcaption{xPatch}
    \end{subfigure}
    
    \caption{Forecasting performance and statistical test under the input-192-predict-96 setting based on 5 select state-of-the-art deep time series models, which is in the vertical layout.}
    \label{fig:acclaw_96_96_vertical}
\end{figure}

\section{Generality Analysis}\label{appdix:general_seq}
\begin{figure}[h]
    \centering
    \begin{subfigure}[b]{\linewidth} 
        \centering
        \includegraphics[width=\linewidth]{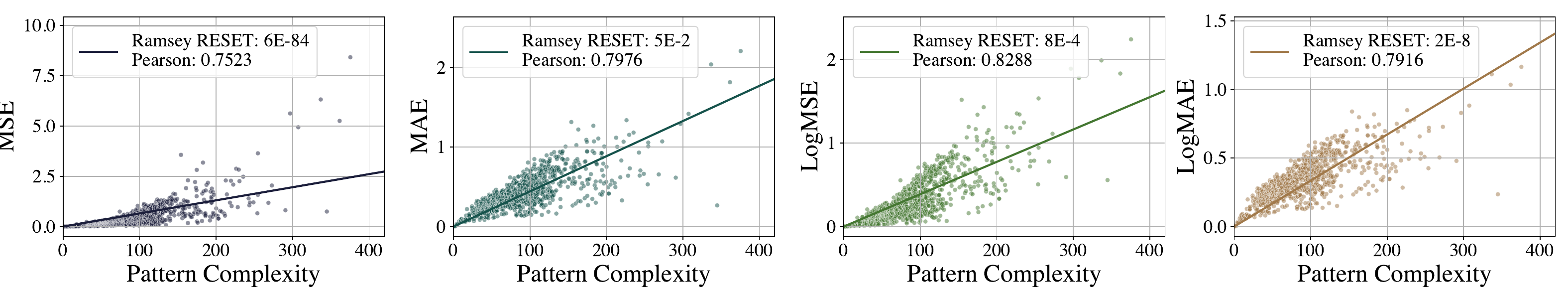} 
        \subcaption{Experimental results under the input-192-predict-96 setting. The fitted coefficient $\alpha=0.0039$}  
    \end{subfigure}
    
    \begin{subfigure}[b]{\linewidth}
        \centering
        \includegraphics[width=\linewidth]{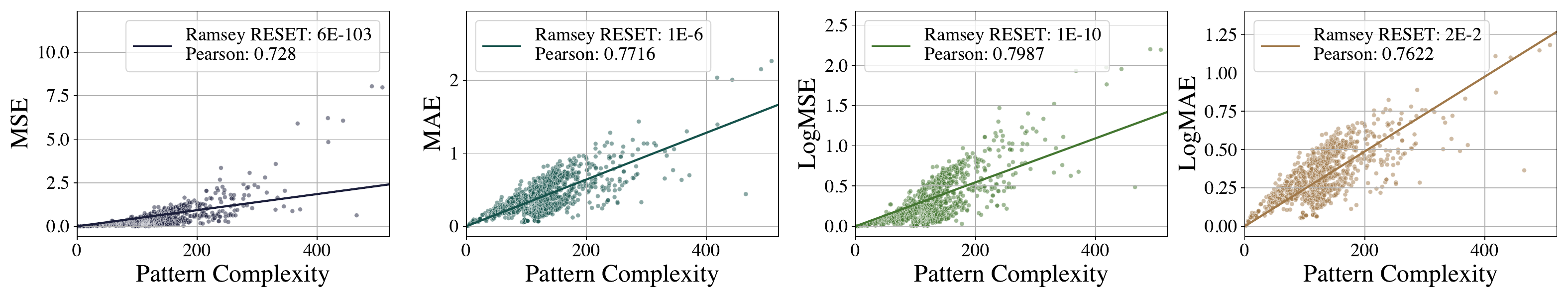} 
        \subcaption{Experimental results under the input-336-predict-96 setting. The fitted coefficient $\alpha=0.027$}
    \end{subfigure}

    \begin{subfigure}[b]{\linewidth}
        \centering
        \includegraphics[width=\linewidth]{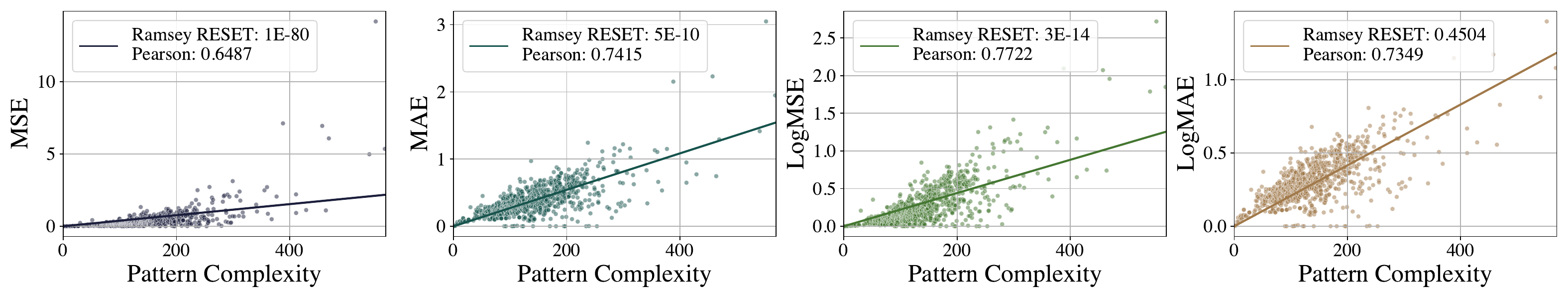} 
        \subcaption{Experimental results under the input-512-predict-96 setting. The fitted coefficient $\alpha=0.0022$}
    \end{subfigure}
    
    \caption{Fitted accuracy law with increased look-back length in $\{192, 336, 512\}$.}
    \label{fig:acclaw_increase_input}
\end{figure}

\begin{figure}[h]
    \centering
    \begin{subfigure}[b]{\linewidth} 
        \centering
        \includegraphics[width=\linewidth]{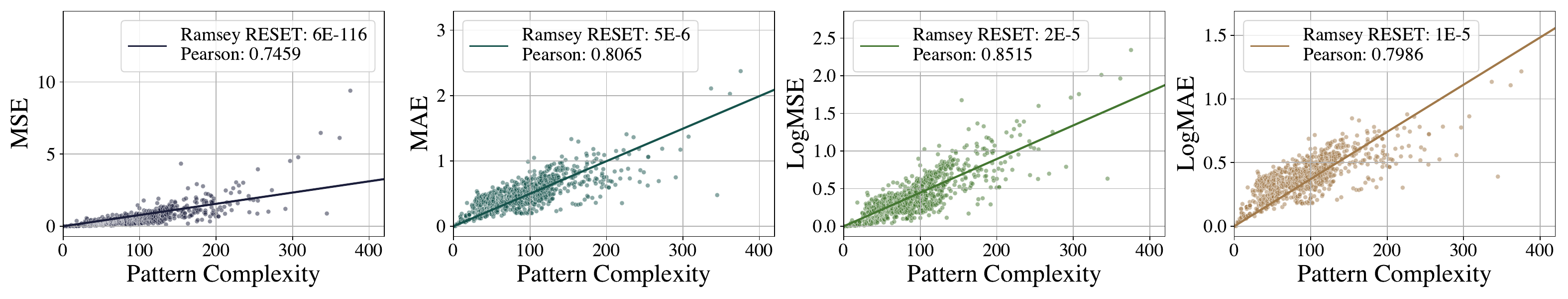} 
        \subcaption{Experimental results under the input-96-predict-192 setting. The fitted coefficient $\alpha=0.0045$}  
    \end{subfigure}
    
    \begin{subfigure}[b]{\linewidth}
        \centering
        \includegraphics[width=\linewidth]{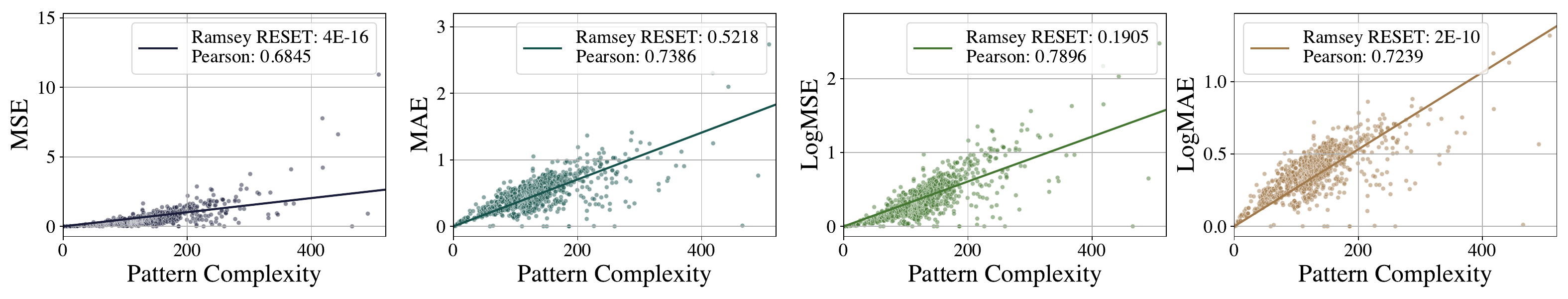} 
        \subcaption{Experimental results under the input-96-predict-336 setting.  The fitted coefficient $\alpha=0.0030$}
    \end{subfigure}
    
    \caption{Fitted accuracy law with increased forecasting length in $\{192, 336\}$.}
    \label{fig:acclaw_increase_output}
\end{figure}

In the main text, we validated the proposed complexity under the input-96–predict-96 setup. To further validate the generality and robustness of our finding, extend the input length and the forecasting horizon, respectively. Figures \ref{fig:acclaw_increase_input}-\ref{fig:acclaw_increase_output} present the experimental results and fitted curves across varying input and output lengths. Notably, the proposed accuracy law, namely, the exponential relationship between series pattern complexity and forecasting performance, is clearly manifest. Notably, as shown in Figure \ref{fig:acclaw_increase_input}, increasing the input length results in a decrease in the estimated coefficient $\alpha$. This aligns with the intuition cause longer input series typically yield better predictions.

\section{Experimental Data Analysis}

\subsection{Data Source}
As described in the main text, the 940 experimental time series is selected from LOTSA \citep{woo2024unified}. Here we list the concrete data source in Table \ref{tab:exp_data}.

\begin{table}[h]
  \centering
  \setlength{\tabcolsep}{10pt}
  \caption{Source of our experimental data. Here we list datasets selected from LOTSA \citep{woo2024unified}, which should contain more than 20 time series and longer than 5000 time steps. Here ``M, H, D'' refers to minutely, hourly and daily sampling frequency respectively.}\label{tab:exp_data}
    \begin{tabular}{lccc}
    \toprule
    \textbf{Dataset} & \textbf{Domain} & \textbf{Frequency} & 
    \textbf{\# Time Series}  \\
    \toprule
    Loop Seattle & Transport & 5M    & 323  \\
    Los-Loop & Transport & 5M    & 207  \\
    PEMS03 & Transport & 5M    & 358   \\
    PEMS04 & Transport & 5M    & 307  \\
    PEMS07 & Transport & 5M    & 883  \\
    PEMS08 & Transport & 5M    & 170  \\ 
    PEMS Bay & Transport & 5M    & 325  \\
    Q-Traffic & Transport & 15M   & 45,148 \\
    LargeST & Transport & 5M    & 42,333 \\
    Traffic Hourly & Transport & H     & 862  \\
    SHMetro & Transport & 15M   & 288 \\
    BDG-2 Panther & Energy & H   & 105 \\
    BDG-2 Fox & Energy & H     & 135  \\
    BDG-2 Rat & Energy & H     & 280  \\
    BDG-2 Bear & Energy & H     & 91  \\
    Buildings900K & Energy & H     & 1,792,328 \\
    BDG-2 Bull & Energy & H     & 41 \\
    BDG-2 Hog & Energy & H     & 24  \\
    KDD Cup 2018 & Energy & H     & 270 \\
    KDD Cup 2022   & Energy & 10M   & 134 \\
    Low Carbon London & Energy & H     & 713 \\
    London Smart Meters & Energy & 30M & 5,520  \\
    Residential Load Power & Energy & M     & 271 \\
    Residential PV Powe  & Energy & M     & 233 \\
    GEF12 & Energy & H     & 20   \\
    Wind Farms & Energy & M     & 337  \\
    CMIP6-2000 & Climate & 6H    & 1,351,680  \\
    CMIP6-2005 & Climate & 6H    & 1,351,680  \\
    CMIP6-2010 & Climate & 6H    & 1,351,680  \\
    ERA5-1997  & Climate & H     & 245,760 \\
    ERA5-1998  & Climate & H     & 245,760 \\
    ERA5-1999  & Climate & H     & 245,760 \\
    ERA5-2000  & Climate & H     & 245,760 \\
    Subseasonal & Climate & D     & 862   \\
    Subseasonal Precipitation & Climate & D  & 862  \\
    \bottomrule
    \end{tabular}
  \label{tab:dataset_desp}%
\end{table}

\subsection{Visualization}
\begin{figure}[h]
    \centering
    \begin{subfigure}[b]{\linewidth} 
        \centering
        \includegraphics[width=0.7\linewidth]{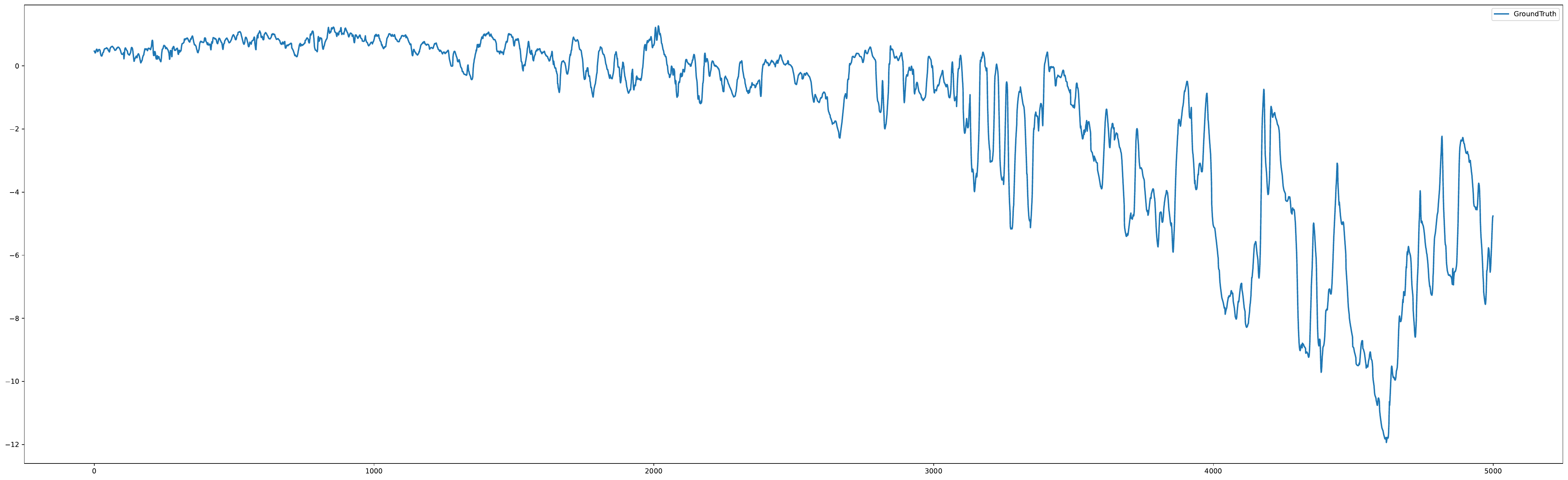} 
        \subcaption{Series from ERA5 2000 with Complexity=300.800}  
    \end{subfigure}

    \begin{subfigure}[b]{\linewidth}
        \centering
        \includegraphics[width=0.7\linewidth]{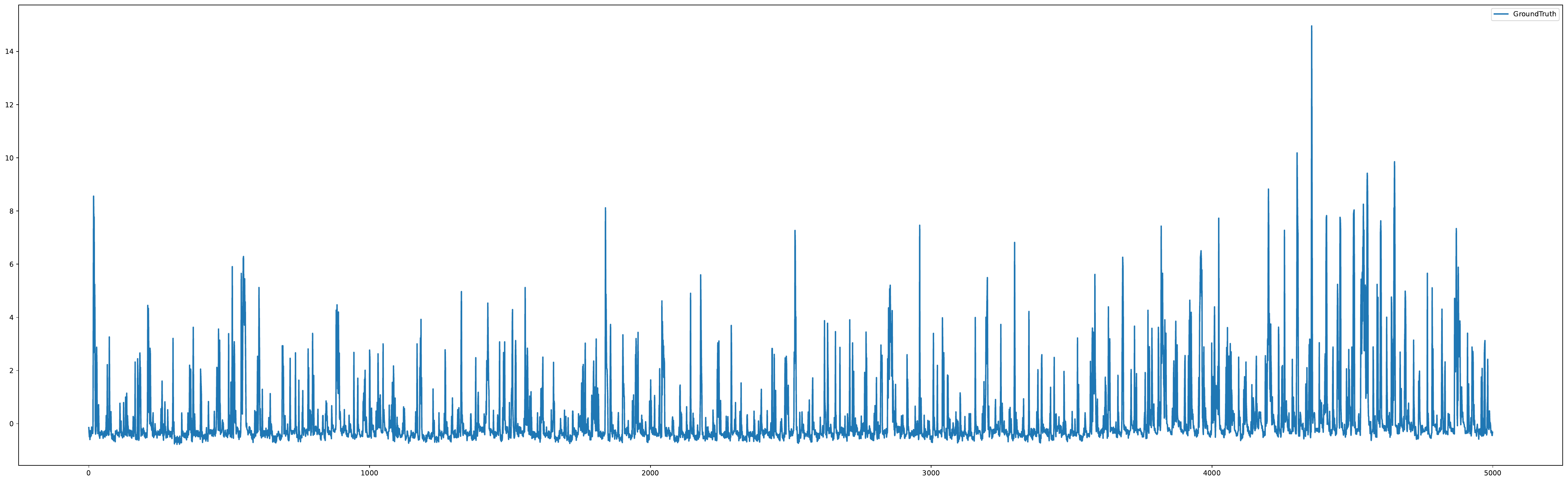} 
        \subcaption{Series from London Smart Meters with Complexity=181.172}
    \end{subfigure}

    \begin{subfigure}[b]{\linewidth}
        \centering
        \includegraphics[width=0.7\linewidth]{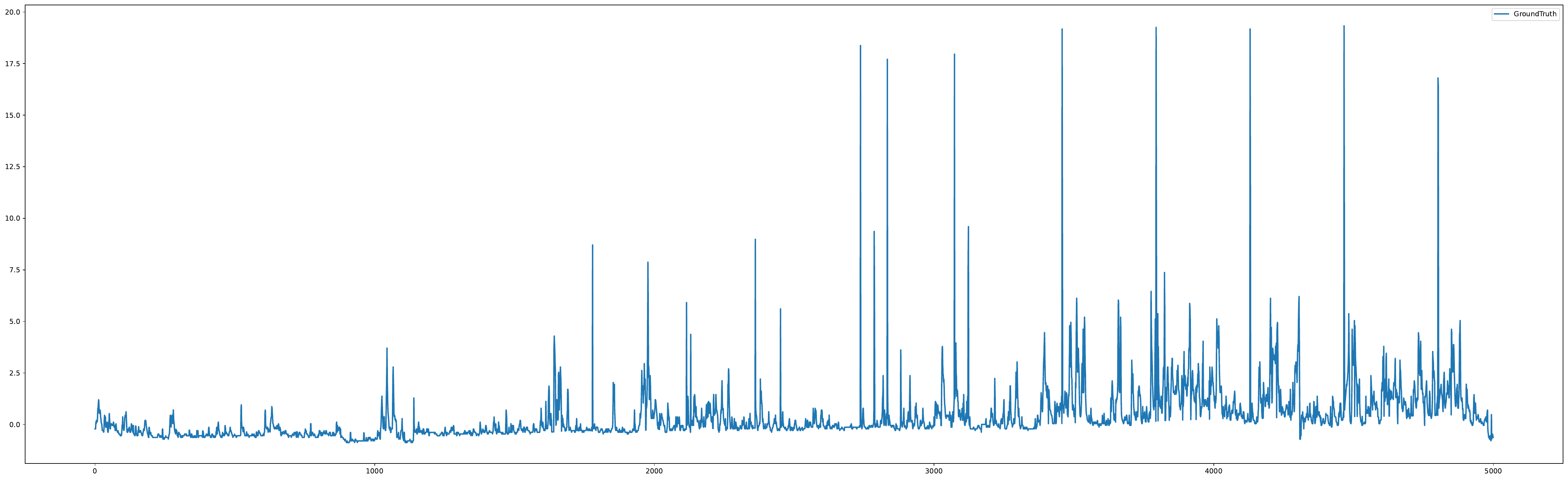} 
        \subcaption{Series from  China Air Quality with Complexity=140.627}
    \end{subfigure}

    \begin{subfigure}[b]{\linewidth}
        \centering
        \includegraphics[width=0.7\linewidth]{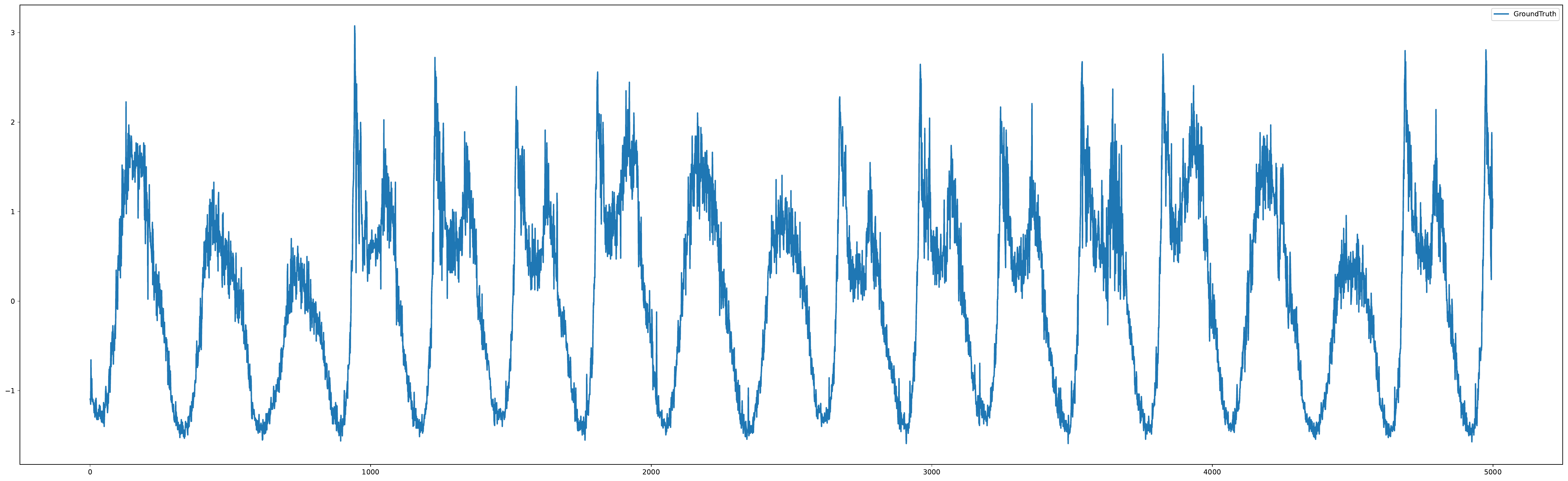}  
        \subcaption{Series from  PEMS03 with Complexity = 68.413}
    \end{subfigure}

    \begin{subfigure}[b]{\linewidth}
        \centering
        \includegraphics[width=0.7\linewidth]{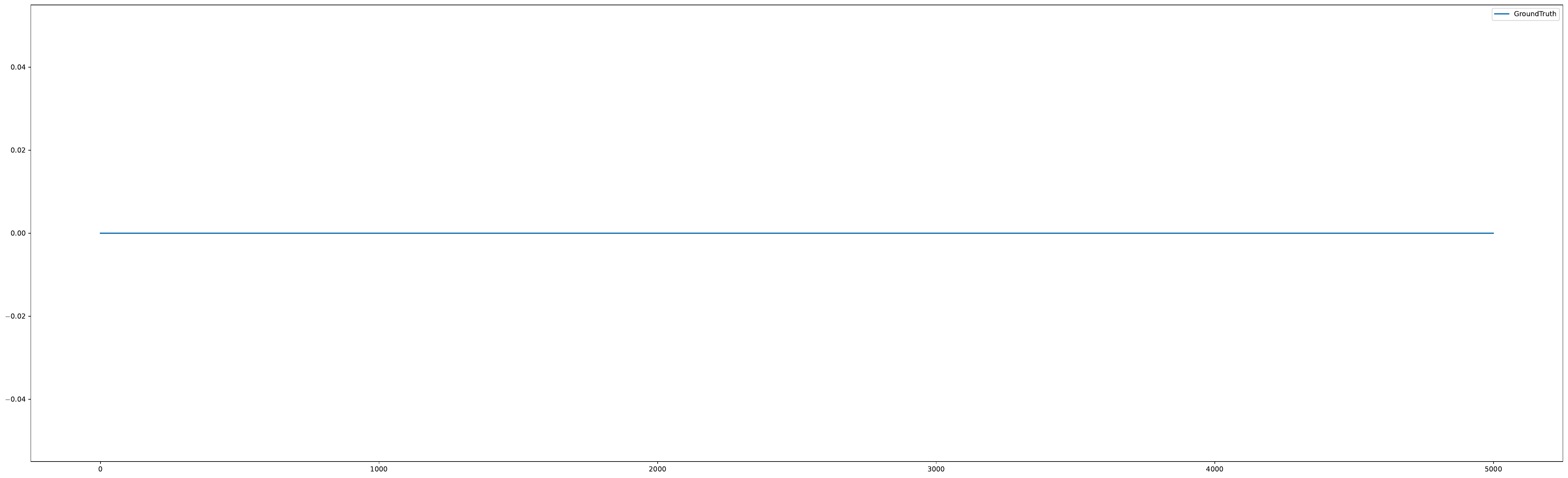}
        \subcaption{Series Residential Load Power with Complexity = 0}
    \end{subfigure}
    
    \caption{Visualizations of time series examples with different pattern complexity values.}
    \label{fig:acclaw_vis}
\end{figure}

Figure \ref{fig:acclaw_vis} illustrates the visualizations of time series data corresponding to varying levels of complexity. It is notable that as complexity decreases, the data becomes simple to learn. When the complexity reaches zero, the data reduces to a straight line, which aligns with intuitive expectations.

\end{document}